\RequirePackage{fix-cm}
\documentclass[twocolumn]{svjour3}          
\smartqed  
\usepackage{graphicx}
\usepackage{amsmath}
\usepackage{amssymb}
\usepackage[utf8]{inputenc}  
\usepackage[T1]{fontenc}  
\usepackage{enumitem}
\usepackage[disable]
	{todonotes}
\newcommand{\todoVL}[1]{\todo[color=red!40, author=VL, inline]{#1}}

\usepackage[bookmarks=true, bookmarksnumbered=true, bookmarksopen=true, bookmarksdepth=3
		unicode=true, colorlinks=true,
		linkcolor=blue,citecolor=blue,filecolor=blue,urlcolor=blue,
 		pdfstartview=FitH
		]{hyperref}

\begin{document}

\title{A Review of Features for the Discrimination of Twitter Users: Application to the Prediction of Offline Influence}


\author{Jean-Valère Cossu \and
        Vincent Labatut \and
        Nicolas Dugué
}


\institute{Jean-Valère Cossu, Vincent Labatut \at
              Université d'Avignon, LIA EA 4128, France \\
              \email{\{jean-valere.cossu, vincent.labatut\}@univ-avignon.fr}           
           \and
           Nicolas Dugué \at
              Université d'Orléans, INSA Centre Val de Loire, LIFO EA 4022, France\\
              \email{nicolas.dugue@univ-orleans.fr}
}

\date{Received: date / Accepted: date}

\maketitle

\begin{abstract}
Many works related to Twitter aim at characterizing its users in some way: role on the service (spammers, bots, organizations, etc.), nature of the user (socio-professional category, age, etc.), topics of interest, and others. However, for a given user classification problem, it is very difficult to select a set of appropriate features, because the many features described in the literature are very heterogeneous, with name overlaps and collisions, and numerous very close variants. In this article, we review a wide range of such features. In order to present a clear state-of-the-art description, we unify their names, definitions and relationships, and we propose a new, neutral, typology. We then illustrate the interest of our review by applying a selection of these features to the offline influence detection problem. This task consists in identifying users which are influential \textit{in real-life}, based on their Twitter account and related data. We show that most features deemed efficient to predict \textit{online} influence, such as the numbers of retweets and followers, are not relevant to this problem. However, we propose several content-based approaches to label Twitter users as \textit{Influencers} or not. We also rank them according to a predicted influence level. Our proposals are evaluated over the CLEF RepLab 2014 dataset, and outmatch state-of-the-art methods.
\end{abstract}

\sloppy
\section{Introduction}
Social Networking Services have started to appear on the World Wide Web as early as the year 2000, with sites such as Friendster and MySpace. Since then, they have multiplied and taken over the Internet, with hundreds of different services used by more than one billion people. Among them, Twitter is one of the most popular. It is used to report live events, share viewpoints regarding a variety of topics, monitor public opinion, track e-reputation, etc. The service consequently dragged the attention of politicians, firms, celebrities and marketing specialists. which now largely base their communication on Twitter, trying to become as \textit{visible} and \textit{influential} as possible.


\textbf{User Classification.} Due to the popularity and widespread use of Twitter, there are numerous reasons why one would want to categorize its users: market segmentation and marketing target identification, detection of opinion trends, quality of service improvement (e.g. by blocking spammers), sociological studies, and others. But because of the diversity of Twitter users and of the amount of available data, there are many ways to do so. For these reasons, many works have been dedicated to the characterization of Twitter profiles.

A number of these studies aim at identifying users holding certain roles inside the service itself. The detection of \textit{spammers} is very popular, due to the critical nature of this task regarding the quality of service. Most works focus on the identification of \textit{spambots}, i.e. software agents working in an automated way \cite{Benevenuto2010,Ghosh2012,Lee2010,Lee2011,Wang2010}. The detection of \textit{crowdturfers}, the human crowdsourced equivalent of spambots, constitutes a related but less-known task \cite{Lee2013}. The tool described in \cite{Chu2012} distinguishes regular human users, \textit{bots} (robots, i.e. fully automated users, which can be spammers, but not necessarily) and so-called \textit{cyborgs} (computer-assisted humans or human-assisted bots).
Certain authors study \textit{social capitalists}, a class of users taking advantage of specific strategies to gain visibility on Twitter without producing any valuable content. Some works focus on their identification \cite{Dugue2014,Dugue2015,Ghosh2012}, others on the characterization of their position and role in the network \cite{Dugue2014a}.
Some typologies are more detailed, for instance in \cite{Uddin2014}, the authors distinguish $3$ types of \textit{real users} (personal, professional, business) and $3$ types of \textit{digital actors} (Spammers, Newsfeeds, Marketing services).
In \cite{Lee2014}, the authors propose a method to detect \textit{Retweeters}, i.e. users more likely to fetch tweets related to a given subject.
Influence is also a topic of interest, with numerous works aiming at measuring it, or detecting influential users \cite{Anger2011,Cossu2015,Weng2010}.

Other works categorize users relatively to real-world aspects. Many works focus on socio-professional categories: age \cite{AlZamal2012,Rangel2014,Rao2010}, gender \cite{AlZamal2012,Rangel2014,Rao2010}, ethnicity/regional origin \cite{Pennacchiotti2011,Rao2010}, city \cite{Cheng2010,Mahmud2012}, country \cite{Huang2014,Mahmud2012}, political orientation \cite{GayoAvello,AlZamal2012,Conover2011,Makazhanov2013,Pennacchiotti2011,Rao2010}, business domain \cite{Pennacchiotti2011}. In \cite{Silva2014}, the authors distinguish two types of Twitter users (individual persons vs. organizations), and three in \cite{Choudhury2012} (organizations, journalists, ordinary persons). 

Certain works categorize users not relatively to the whole system, but to some user of interest. This is noticeably the case in works aiming at recommending followees (i.e. which users to follow) \cite{armentano2011topology,golder2009structural,garcia2010weighted,kywe2012survey}.
Some works aim at simultaneously classifying users according to topics/categories not specified in advance, and uncover the most relevant topic/categories themselves \cite{Java2007,Ramage2010}.
Finally, another category of works takes advantage of user-related features to improve the classification of tweets. For instance, several articles describe methods to distinguish tweets depending on the communication objective behind them. In \cite{Sriram2010}, the authors distinguish \textit{News}, \textit{Opinions}, \textit{Deals}, \textit{Events} and \textit{Private messages} ; in \cite{Naaman2010} they use $9$ categories such as \textit{Information sharing}, \textit{Self promotion}, and \textit{Question to followers}.

\textbf{Twitter Features.} The cited studies come from a variety of fields: computer science, sociology, statistical physics, political sciences, etc. They consequently have different goals, and tackle the problem of user classification in different ways, applying different methods to different data. However, the adopted approaches can be commonly described in a generic way: 1) identifying the appropriate features, i.e. the relevant data describing the users ; and 2) training a classifier to discriminate the targeted user classes based on these features. In this article, we focus on the first point, i.e. the features one can extract from Twitter for the purpose of user classification. 

Over the years, and because user classification studies come from such a variety of backgrounds, a number of such features have been proposed for the purpose of user classification. Some are specific to certain research domains. For instance, works coming from Social Network Analysis (SNA) tend to focus on the way users are interconnected, whereas studies from Natural Language Processing (NLP) obviously focus on the textual content of tweets. But many simple features, such as the number of Tweets published by a user, are widespread independently from the research domain. The difficulty for a newcomer is that, over those articles, these features may have different names when they actually are the same ; or \textit{vice versa} (same name when they actually are different) ; or one feature can be declined into a number of more or less similar variants. Moreover, it is difficult to determine which feature or variant is appropriate for a given user classification problem: the features one would use to detect spammers might not be relevant when trying to identify the political orientation of users. For instance, during the 3rd International Author Profiling Task at PAN 2015 \cite{Rangel2015}, which focused on Age and Gender identification, the organizers were not able to highlight a specific, particularly relevant feature.

\textbf{Contributions.} In this article, we propose a review of the features used to classify Twitter users. Of course, performing an exhaustive survey seems hardly possible, due to the number of concerned works. We however consider a wide range of studies and adopt a high level approach, focusing on the meaning of the features while also describing the different forms they can take. We organize them in a new, trans-domain typology. As an illustration of how our review can be used, we then apply a selection of these features to a real-world problem: the detection of \textit{offline influential users}. In other words, we aim to solve the problem consisting in detecting people which are influential \textit{in real-life}, based on their Twitter profile and activity. To answer this question, we conduct experiments on the CLEF RepLab 2014 dataset, which was designed specifically for this task. Indeed, it contains Twitter data including Twitter profiles annotated in terms of offline influence. We take advantage of these manual annotations to train several Machine Learning (ML) tools and assess their performance on classification and ranking issues. The former consists in determining if a user is influential or non-influential, whereas the latter aims at ranking users depending on their estimated influence level.

Our first contribution is to review a large number of Twitter-based features used for user profile characterization problems, and to present them in a unified form, using a new typology. Our second contribution is the assessment of these generic features, relatively to a specific problem consisting in predicting offline influence. We show that most simple features behave rather poorly, and discuss the questions raised by this observation. Finally, we describe several NLP ranking methods that gives better results than known state-of-the-art approaches.

\textbf{Organization.} The rest of this paper is organized as follows. In the next section (Section \ref{sec:Features}), we review the features related to the classification or characterization of Twitter users, with an emphasis on their meaning in this context. We also propose a typology for these features, in an effort to highlights how they are connected. We then focus on Twitter-based \textit{offline influence detection} in Section \ref{sec:Application}. We describe the problem at hand, the RepLab data we used in our experiments, and the methods we propose to solve this problem. In Section \ref{sec:Results}, we present the results we obtained and discuss them. Finally, we highlight the main aspects of our work in the conclusion, and give some perspectives regarding how it can be extended.

\section{Review of Twitter-Related Features}
\label{sec:Features}
We present a review of the most interesting features one can use to characterize Twitter users. Due to the generally large number of features used in a given study, authors often group them thematically. However, there is no standard regarding the resulting feature categories, which can vary widely from one author to the other. In particular, people tend to categorize features based on some criteria related to their field of study (i.e. mainly SNA and NLP). Here, we try to ignore this somehow artificial distinction, and propose a neutral typology. We do not want to be exhaustive, but rather to include widely used features, and to emphasize their diversity.

Before starting to describe the features in detail, we need to introduce some concepts related to Twitter. This online \emph{micro-blogging} service allows to publicly discuss largely publicized as well as everyday-life events \cite{Java2007} by using \emph{tweets}, short messages of at most $140$ characters. To be able to see the tweets posted by other users, one has to \textit{subscribe} to these users. If user $u$ subscribes to user $v$, then $u$ is called a \textit{follower} of $v$, whereas $v$ is a \textit{followee} of $u$. Each user can \textit{retweet} other users' tweets to share these tweets with his followers, or mark his agreement \cite{Boyd2010}. Users can also explicitly \textit{mention} other users to drag their attention by adding an expression of the form \texttt{@UserName} in their tweets. One can reply to a user when he is mentioned. Another important Twitter feature is the possibility to tag tweets with key words called \textit{hashtags}, which are strings marked by a leading sharp (\#) character.

Table \ref{tab:Features} presents the list of all the features we reviewed, indicating for each one: its category, a short description of the feature, one or several associated descriptors (i.e. values representing the feature), and some bibliographic references illustrating how the feature was used, when possible. Sometimes, several descriptors are indicated for the same feature, because it can be used in various ways. This is particularly true for those which can be expressed as a value for each tweet, for example the number of mentions in a tweet (Feature \ref{feat:OutMentions}). It is possible to treat them in an absolute way, i.e. sum of the values over the considered period (e.g. total number of mentions) or keep only the extreme values (e.g. minimal and maximal numbers of mentions). One can also use a relative approach by processing a central and a dispersion statistics (e.g. average number of mentions by tweet, and the corresponding standard deviation). 

\begin{table*}[!t]
	\renewcommand{\arraystretch}{1.3}
	\newcounter{FeatureList}
	\newcommand{\myItem}[1]{\refstepcounter{FeatureList}\label{#1}\vspace{-0.22cm}\arabic{FeatureList}.}
	\caption{Features used to characterize Twitter users. The \textit{Descriptor} column indicates which statistics can be used to represent the feature: Total count (\textit{Cnt}), Average value (\textit{Avg}), Standard deviation (\textit{Sd}), Minimum value (\textit{Min}), Maximum value (\textit{Max}), Overall proportion (Prop), Set cardinality (\textit{Cardinality}). The number of examples is limited to $5$.}
	\label{tab:Features}
	\centering
	\begin{tabular}{p{1.5cm}p{8.1cm}p{3.8cm}p{2.3cm}}
		\hline
		\textbf{Category} & \textbf{Description} & \textbf{Descriptors} & \textbf{Examples}\\
		
        \hline
		User & \myItem{feat:Picture} Profile picture & Boolean/Image & \cite{Pennacchiotti2011,Vilares2014} \\
        Profile & \myItem{feat:Verified} Verified account & Boolean & \cite{Chu2012,Lee2010,Uddin2014,Vilares2014} \\
        & \myItem{feat:Contrib} Contributions allowed & Boolean & \cite{Vilares2014} \\
        & \myItem{feat:Webpage} Personal Webpage set & Boolean & \cite{Lee2013,Vilares2014} \\
        & \myItem{feat:DescrLength} Number of characters in the profile description & Count & \cite{Lee2011,Lee2013} \\
        & \myItem{feat:DescrNames} Number of usernames in the profile description & Count & \cite{Ramirez2014} \\
        & \myItem{feat:DescrUrls} Number of URLs in the profile description & Count & \cite{Ramirez2014} \\
        & \myItem{feat:DescrCont} Content of the profile description & Text & \cite{Pennacchiotti2011} \\
        & \myItem{feat:NameLength} Number of (special) characters in the username & Count & \cite{Lee2011,Lee2013,Pennacchiotti2011,Ramirez2014} \\
        & \myItem{feat:ProfileAge} Age of the profile & Value & \cite{Benevenuto2010,Lee2011,Pennacchiotti2011,Ramirez2014,Uddin2014} \\
        & \myItem{feat:Client} Twitter client & Prop/Cnt/Boolean & \cite{Chu2012,Dugue2014,Huang2014} \\
		
       \hline
		Publishing & \myItem{feat:Tweets} Tweets published by the user & Cnt/Avg/Sd/Min/Max & \cite{Chu2012,Lee2011,Ramirez2014,Rao2010,Vilares2014} \\
		Activity & \myItem{feat:Medias} Media resources published by the user & Cnt/Prop/Avg/Sd/Min/Max & \cite{Ramirez2014} \\
		& \myItem{feat:Delay} Delay between two consecutive tweets of the user & Avg/Sd/Min/Max & \cite{Benevenuto2010,Pennacchiotti2011,Ramirez2014} \\
        & \myItem{feat:SelfMentions} Self-mentions of the user & Cnt/Prop/Avg/Sd/Min/Max & \cite{Ramirez2014} \\
        & \myItem{feat:Geolocated} Geolocated tweets published by the user & Prop/Cnt/Boolean & \cite{Huang2014,Vilares2014} \\
		
        \hline
		Local & \myItem{feat:NbrFollow} Topology of the follower-followee network & Graph-related measures & \cite{Cha2010,Lee2011,Ramirez2014,Tommasel2015,Vilares2014} \\
        Connections & \myItem{feat:NbrLists} Subscription lists containing the user & Count & \cite{Danisch2014,Vilares2014}\\
        & \myItem{feat:IdFollow} Ids of the user's most recent followers/followees & Standard deviation & \cite{Lee2011} \\
        & \myItem{feat:TweetFollow} Tweets published by the followers/followees & Cnt/Avg/Sd/Min/Max & \cite{Benevenuto2010,Ramirez2014} \\
		
        \hline
		User & \myItem{feat:OutRetweets} Retweets published by the user & Cnt/Prop/Avg/Sd/Min/Max & \cite{Benevenuto2010,Danisch2014,Pennacchiotti2011,Rao2010,Uddin2014} \\
        Interaction & \myItem{feat:InRetweets} Number of times the user is retweeted by others & Cnt/Prop/Avg/Sd/Min/Max & \cite{Anger2011,Benevenuto2010,Cha2010,Ramirez2014} \\
        & \myItem{feat:OutFavs} Favorites selected by the user & Count & \cite{Choudhury2012,Ramirez2014,Vilares2014} \\
        & \myItem{feat:InFavs} Tweets of the user marked as favorite by others & Cnt/Prop/Avg/Sd/Min/Max & \cite{Danisch2014,Ramirez2014,Uddin2014} \\
        & \myItem{feat:OutMentions} (Unique) mentions \textit{of} other users & Cnt/Prop/Avg/Sd/Min/Max & \cite{Chu2012,Lee2011,Ramirez2014,Silva2014,Uddin2014} \\
        & \myItem{feat:InMentions} Mentions \textit{by} other users & Cnt/Avg/Sd/Min/Max & \cite{Benevenuto2010,Cha2010,Uddin2014} \\
		
        \hline
		Lexical & \myItem{feat:Words} Number of (unique) words & Cnt/Avg/Sd/Min/Max & \cite{Benevenuto2010,Ramirez2014,Weren2014} \\
        Aspects & \myItem{feat:Hapaxes} Number of hapaxes & Cnt/Prop/Avg/Sd/Min/Max & \cite{Ramirez2014} \\
        & \myItem{feat:NamedEnt} Named entities & Cnt/Prop/Avg/Sd/Min/Max & \cite{Choudhury2012,Silva2014} \\
        & \myItem{feat:NgramWeights} Word $n$-gram weighting & Vector & \cite{Conover2011,Cossu2015,Silva2014,Vilares2014,Weren2014} \\
        & \myItem{feat:ProtoWords} Prototypical $n$-grams & Vector & \cite{AlZamal2012,Cheng2010,Lee2013,Makazhanov2013,Pennacchiotti2011} \\

		\hline
		Stylistic & \myItem{feat:WordLength} Word length, in characters & Avg/Sd/Min/Max & \cite{Ramirez2014} \\
        Traits & \myItem{feat:TweetLength} Tweet length & Avg/Sd/Min/Max & \cite{Benevenuto2010,Makazhanov2013,Ramirez2014,Silva2014} \\
        & \myItem{feat:TweetRead} Readability of the user's tweets & Avg/Sd/Min/Max & \cite{Silva2014,Weren2014} \\
        & \myItem{feat:SpecialChars} Special characters or patterns & Cnt/Prop/Avg/Sd/Min/Max & \cite{Benevenuto2010,Ramirez2014,Rao2010,Silva2014,Weren2014} \\
        & \myItem{feat:Hashtags} Number of (unique) hashtags & Cnt/Prop/Avg/Sd/Min/Max & \cite{Benevenuto2010,Lee2011,Pennacchiotti2011,Silva2014,Uddin2014} \\
        & \myItem{feat:Urls} Number of (unique) URLs & Cnt/Prop/Avg/Sd/Min/Max & \cite{Chu2012,Lee2010,Pennacchiotti2011,Ramirez2014,Uddin2014} \\
        & \myItem{feat:AutoSimilarity} Similarity between the user's own tweets & Cnt/Avg/Sd/Min/Max & \cite{Lee2011,Lee2013,Wang2010} \\
		
        \hline
        External & \myItem{feat:WebSearch} Number of Web search results for the user's page & Count & \cite{Cossu2015} \\
		Data & \myItem{feat:Klout} Klout score & Value & \cite{Cossu2015,Dugue2015} \\
        & \myItem{feat:Kred} Kred score & Value & \cite{Dugue2015} \\

		\hline        
	\end{tabular}
\end{table*}

\subsection{Description of the Features}
This subsection describes all the features from Table \ref{tab:Features} in detail, considering each category separately. We discuss each feature and indicate how it is relevant, and in which context.

\subsubsection{User Profile}
Our first category gathers features related to user profiles. The first 4 are Boolean values representing whether: the user set up a profile picture (Feature \ref{feat:Picture}), his account was officially verified by Twitter (Feature \ref{feat:Verified}), he allows other users to contribute to his account (Feature \ref{feat:Contrib}), 
he set up his personal Webpage (Feature \ref{feat:Webpage}). The profile picture itself is also analyzed by certain authors, using image processing methods, to extract information such as age, gender and race \cite{Huang2014}.

Feature \ref{feat:DescrLength} is an integer corresponding to the length (in characters) of the text the user wrote to describe himself. These features are good indicators of how committed the user is regarding Twitter and his online presence. Professional bloggers and corporate accounts, in particular, generally fill these profile fields, whereas spambots, or passive users (i.e. only reading Twitter feeds but not producing any content) do not. Verified accounts tend to be owned by humans, not bots \cite{Chu2012}.

The content of the profile description can also be analyzed (Feature \ref{feat:DescrCont}) in order to extract valuable information. For instance, in \cite{Pennacchiotti2011}, Pennacchiotti \& Popescu engineered a collection of regular expressions in order to retrieve the age and ethnicity of the users.

Features \ref{feat:DescrNames} and \ref{feat:DescrUrls} are the number of usernames and URLs appearing in the textual profile description. Indeed, certain users take advantage of this text to indicate they have other accounts or reference several Websites. This can concern users with several professional roles they want to distinguish, as well as users wanting to gain visibility through specific strategies. On the same note, the length of the username (Feature \ref{feat:NameLength}), expressed in characters, was used in some studies to identify certain types of users \cite{Lee2011,Ramirez2014}. For instance, social capitalists tend to have very long names
. Certain authors also focus on the number of special characters (e.g. hearts, emoticons) in the username \cite{Pennacchiotti2011}, which may be characteristic of certain social categories. The name itself can also be a source of information: in \cite{Huang2014}, Huang \textit{et al}. use it to infer the ethnicity of the user.

The age of the profile (Feature \ref{feat:ProfileAge}) is likely to be related to how visible the user is on Twitter, since it takes some time to reach an influential position. It can also help identifying bots: in their 2012 paper, Chu \textit{et al}. noticed $95\%$ of the bots were registered in 2009.

Finally, Feature \ref{feat:Client} corresponds to the software clients the user favors when accessing Twitter: official Web site, official smartphone application, management dashboard tool, third party applications (Vine, Soundcloud...), etc. One can consider each client as a Boolean value representing whether the user regularly takes advantage of this tool \cite{Chu2012,Dugue2015,Huang2014}. Alternatively, it is also possible to select the usage frequency of the tool, expressed in terms of total number or proportion of uses.

\subsubsection{Publishing Activity}
The next category focuses on the way the user behaves regarding the tweets he publishes. Feature \ref{feat:Tweets} corresponds to the number of tweets he posted on the considered period of time, so it represents how active the user is globally. Users posting a small number of tweets are potentially information seekers \cite{Java2007}. Because this number is generally high, certain authors prefer to consider the number of tweets published by day \cite{Lee2011,Pennacchiotti2011}. The standard deviation, minimal or maximal number of tweets published in a day give an idea of the regularity of the user in terms of tweeting. Alternatively, it is also possible to specifically detect periodic posting behaviors, as Chu \textit{et al}. did to identify bots (programs that tweet automatically) \cite{Chu2012}.

Feature \ref{feat:Medias} is the number of media resources contained in these tweets. One can alternatively consider the proportion of the user's tweets containing a media resource, or one of the previously cited statistics for a given period of time (e.g. by day). The fact a user posts a lot of pictures or videos could be discriminant in certain situations. For instance, the concerned user could be active in an image-related field such as photography, or he could tweet professionally to advertise for a company. 

Feature \ref{feat:Delay} is the duration between two consecutive tweets. It aims at representing how regularly the user tweets. Authors generally focus on the average delay and the associated standard deviation \cite{Pennacchiotti2011}, but the minimum, maximum and median are also used \cite{Benevenuto2010}.

Feature \ref{feat:SelfMentions} is the number of mentions the user makes of himself. This strategy is used by users who need several consecutive tweets to express an idea, and want to force Twitter to group them in its graphical interface \cite{Greenfield2014}. One can alternatively consider the proportion of the user's tweets containing a self-mention, or the average number of self-mentions by day (or any other statistic listed in Table \ref{tab:Features}, like for the previous features).

Finally, Feature \ref{feat:Geolocated} is the proportion of tweets published by the user which are geolocated. In certain studies, the authors define it instead as a Boolean feature, depending on whether or not the geolocation options is enabled in the user's profile \cite{Vilares2014}. Others prefer to count the number of distinct locations associated to the user \cite{Cossu2015,Huang2014}. Like Features \ref{feat:Picture}--\ref{feat:DescrLength}, this feature can help discriminating certain types of users aiming at exhibiting a very complete and controlled image, or with a specific behavior implying the publicization 
of their physical location (e.g. to draw a crowd in a specific place). In \cite{Huang2014}, the nature of the location is used to identify the user's nationality.

\subsubsection{Local Connections}
The features from this category describe how the user is connected to the rest of the Twitter network. Feature \ref{feat:NbrFollow} corresponds to the network of follower-to-followee relationships, which can be treated in many ways. Most authors extract two distinct values to represent a user: the number of followers (people which have subscribed to the user's feed) and the number of followees (people to which the users have subscribed). In other words, the incoming and outgoing degrees of the node representing the user in the network, respectively.

Some authors alternatively consider the set obtained by taking the \textit{intersection} of the user's followers and followees. For instance, Dugué and Perez \cite{Dugue2014} used it to distinguish regular users from so-called \textit{social capitalists}. These particular users take advantage of specific strategies allowing them to be highly visible on Twitter, while producing absolutely no content of interest. One of the consequences of this behavior is a strong overlap between their followers and followees, which can be identified through the mentioned intersection. Furthermore, this descriptor was used in Golder et al. for followee recommendation \cite{golder2009structural}. More generally, the friends and followers sets are commonly used by recommender systems to model the user interests \cite{garcia2010weighted,armentano2011topology}. 
Also note that a number of combinations of these set-based values appear in the literature. Such combinations are specifically treated in Section \ref{sec:FeatGenRemarks}, but the follower-to-followee ratio is worth mentioning, since is the most widespread \cite{AlZamal2012,Benevenuto2010,Lee2013,Rao2010,Wang2010,garcia2010weighted}.

Some other authors prefer to use the network in a more global way, instead of focusing only on the local topology. For instance, Weng \textit{et al}. \cite{Weng2010} proposed a modification of the PageRank algorithm which allows to compute an influence score for a given topic. Java \textit{et al}. used the HITS centralities (hub and authority measures) to detect users of interest, and community detection to identify groups of users concerned by the same topics \cite{Java2007}. However, these methods require to obtain the whole network, which is generally hardly possible.

Subscription lists allow Twitter users to group their followees as they see fit, and to share these lists with others. Placing a user in such a list can consequently be considered as a stronger form of subscription. Certain authors thus use the number of lists to which a user belongs as such a feature (Feature \ref{feat:NbrLists}).

Like Feature \ref{feat:NbrFollow}, Feature \ref{feat:IdFollow} is dual, in the sense it can be processed for followers and for followees. It is the standard deviation of the ids of the people who recently subscribed to the user's feed, or of the people to which the user recently subscribed. Spambot farms tend to create numerous fake accounts and make them subscribe to each other, in order to artificially increase their visibility. The fake accounts are created rapidly, so the associated numerical ids tend to be near-consecutive: this can be detected by Feature \ref{feat:IdFollow}.

Finally, Feature \ref{feat:TweetFollow} is also dual, it is the numbers of tweets published by the user's followers and by his followees. It represents the level of publishing activity in the direct neighborhood of the user of interest. Instead of a raw count, one can alternatively average by neighbor, or use one of the other statistics listed in Table \ref{tab:Features}. Like for Feature \ref{feat:Tweets}, it is also possible to consider a time period, e.g. the average number of tweets published by the user's followers by day.

\subsubsection{User Interaction}
This category gathers features describing how the user and the other people interact. Feature \ref{feat:OutRetweets} is the proportion of retweets among the tweets published by the user, i.e. the proportion of other persons' messages that the user relayed \cite{Choudhury2012,Benevenuto2010,Rao2010}. It is also possible to consider the raw count of such retweets \cite{Uddin2014}, or to process a time-dependent statistic such as the average number (or proportion) of retweets by day. Symmetrically, Feature \ref{feat:InRetweets} is the number of times a tweet published by the user was retweeted by others. Alternatively, one can also use the proportion of the user's tweets which were retweeted at least once \cite{Anger2011}. These features represent how much the user reacts to external tweets, and how much reaction he gets from his own tweets. Alternatively, certain authors worked with the retweet network, i.e. a graph in which nodes represent users and are connected when one user retweets another. In \cite{Conover2011}, Conover \textit{et al}. applied a community detection algorithm to this network, in order to extract a categorical feature (the community to which a user belongs).

Features \ref{feat:OutFavs} and \ref{feat:InFavs} are related to the ability Twitter users have to mark certain tweets as their favorites. Feature \ref{feat:OutFavs} is the total number of favorites selected by the user, whereas Feature \ref{feat:InFavs} is the number of times a tweet published by the user was marked as favorite by others. Considering an average value by day is not really relevant for the former, because the number of favorites is generally small. However, this (or another statistic) might be more appropriate for the latter, since the number is likely to be higher. Like the previous ones (Features \ref{feat:OutRetweets} and \ref{feat:InRetweets}), these features are related to the reactions caused by tweets. However, a retweet is a much easier and frequent operation, which gives more importance to favorites.

The two last features deal with mentions, i.e. the fact of explicitly naming a user in a tweet. Feature \ref{feat:OutMentions} is the number of mentions the user puts in his tweets. Certain authors count only \textit{unique} mentions (i.e. they do not count the same mention twice), whereas others consider all occurrences. This feature allows identifying the propensity a user has to directly converse with other users. Spambots are also known to fill their tweets with many more mentions than human users \cite{Wang2010}. Instead of counting the mentions, certain authors use their length. Indeed, as we have seen for Feature \ref{feat:NameLength}, the length of a username (mentions are based on usernames) can convey a relevant information. It is also possible to compute the proportion of the user's tweets which contain mentions to other users.

Feature \ref{feat:InMentions} is symmetrical to Feature \ref{feat:OutMentions}: it is the number of times the user is mentioned by others. It can be averaged (or any other statistic) for a given period of time (e.g. number of mentions by day). It can also be divided by the number of tweets published by the user, to get an average number of answers by user's tweet (mentions generally express the will to answer another user's tweet). This feature is interesting, but computationally hard to process, because for a given user, it basically requires parsing all tweets published by the other users. So, it is treatable only for small datasets.

\subsubsection{Lexical Aspects}
This category deals with the content produced by the user. A number of features can be used to describe the lexical aspects of the text composing his tweets. They are relevant when one wants to discriminate users depending on the ideas they express on Twitter, or how they express them. For instance, if a class of users tend to tweet about the same topic, these features are likely to allow their identification.

Feature \ref{feat:Words} is related to the size of the user's lexicon, it is the number of words he uses. It is possible to count all occurrences or to focus only on unique words. Alternatively, one can also compute a statistic expressed by tweet (e.g. average number of unique words by tweet), or over a period of time (e.g. by day). Certain authors prefer to compare the size of the user's lexicon to that of the English dictionary, under the form of a ratio \cite{Weren2014}. Feature \ref{feat:Hapaxes} is very similar, but for \textit{hapaxes}, i.e. words which are unique \textit{to the user} \cite{Ramirez2014}. Put differently, this feature is about words only the considered user includes in his tweets. Instead of counting them, one could also consider the proportion of user's tweets containing at least one hapax. 

Feature \ref{feat:NamedEnt} corresponds to the number of \textit{named entities} identified in the user's tweets. Named entities correspond roughly to proper names, allowing to identify persons, organizations, places, brands, etc. In \cite{Silva2014}, de Silva \& Riloff use the average number of occurrences by tweet, and treat separately each entity type (for persons, organizations and locations). In \cite{Choudhury2012}, de Choudhury \textit{et al}. just consider the absence/presence of entities (i.e. a Boolean feature) in the users's tweets.

Feature \ref{feat:NgramWeights} consists in representing each user by a numerical \textit{vector}. So, it is different from all the other features, which take the form of scalar values (i.e. they represent a user by a \textit{single} value). This feature consequently requires to be processed differently than the others, as illustrated in section \ref{sec:ExperimentalSetup} when treating influence. Feature \ref{feat:NgramWeights} directly comes from the Information Retrieval field \cite{SparckJones1972}. Each value in the vector corresponds to (some function of) the frequency of a specific $n$-gram. In our context, a $n$-gram is a group of $n$ consecutive words. In the simplest case, this value would be the \textit{raw term frequency}, i.e. the number of occurrences of the $n$-gram for the user of interest. However, this frequency can be normalized in different ways (e.g. logarithmic scale), and weighted by quantities such as the \textit{inverse document frequency} (which measures the rarity of the term), resulting in a number of variants. We present a few of them in more details in our application (section \ref{sec:ExperimentalSetup}). 

Many authors use \textit{unigram} weighting (i.e. $1$-grams, or single words) to take advantage of the tweets content, either by itself \cite{Cossu2015} or in combination with other features \cite{Conover2011,Rao2010,Silva2014,Vilares2014,Weren2014}. Other authors also focus on \textit{bigrams} ($2$-grams, or pairs of words) \cite{AlZamal2012,Rao2010,Silva2014,Vilares2014}, for which the same weighting schemes can be applied than for unigrams. But it is also possible to define new ones, for instance by taking advantage of the cooccurrence graphs one can build from bigrams \cite{Cossu2015} (more details on this in Section \ref{sec:ExperimentalSetup}). 

Instead of weights, it is alternatively possible to use $n$-grams to identify the so-called \textit{prototypical expressions} associated to each considered class. One can then characterize a user by looking for these expressions in his tweets. Here, the word \textit{class} is used in a broad sense, and does not necessarily refer to a category of users: certain authors use prototypical words to describe sentiments \cite{Lee2013,Pennacchiotti2011,Silva2014,Weren2014}, or locations \cite{Cheng2010}. Others prefer to focus on topic distillation, i.e. identifying simultaneously some topics and the words that characterize them, and describing users in terms of their interest for these topics depending on their use of the corresponding words \cite{Aleahmad2014,Conover2011,Weng2010}. Moreover, the prototypical expressions correspond to $n$-grams, so certain authors focus on unigrams \cite{AlZamal2012,Choudhury2012,Makazhanov2013,Pennacchiotti2011} while others use bigrams \cite{AlZamal2012} or even trigrams ($3$-grams, or triplets of words) \cite{AlZamal2012,Rao2010}.

\subsubsection{Stylistic Traits}
The tweet content can also be described using non-lexical features, which are gathered in this category. Features \ref{feat:WordLength} and \ref{feat:TweetLength} are the numbers of characters by words, and by tweet, respectively. The length of a tweet is also sometimes expressed in words instead of characters. These features can help characterizing certain types of users. For example, the content tweeted by certain spambots is just a bag of keywords without proper grammatical structure (e.g. \cite{Laasby2014}), which results in a higher average word length.

On the same note, Feature \ref{feat:TweetRead} relies on a measure quantifying the readability of the tweet. This can correspond to the difficulty one would have to understand its meaning \cite{Weren2014}, or to the level of correctness of the text (lexically and/or grammatically) \cite{Silva2014}. For instance, de Silva \& Riloff use the latter to distinguish personal users from companies tweets (which are generally more correct) \cite{Silva2014}.

Feature \ref{feat:SpecialChars} focuses more particularly on special characters, i.e. non-alphanumerical ones, and/or specific patterns such as emoticons and acronyms (\textit{LOL}, \textit{LMFAO}). 
The use of special characters is typical of certain spammers, who substitute some characters to others of the same shape (e.g. $\varepsilon$ for \textsc{E}) in order to convey the same message without being detected by antispam filters. 
Certain authors directly look for emoticons \cite{Rao2010}, which are not used uniformly by all classes of users: according to Rao \textit{et al}., women tend to use them more. Some emoticons can even be processed to identify the sentiment expressed in the tweet \cite{Silva2014}. Other patterns of interest include characters repeated multiple times (e.g. \textit{I am sooooo bored} or \textit{what ?!!!!}) \cite{Silva2014,Weren2014}, pronouns, which are used by de Silva \& Riloff to distinguish individual persons from organizations \cite{Silva2014}, digits \cite{Benevenuto2010}, spam-related words \cite{Benevenuto2010}.

Features \ref{feat:Hashtags} and \ref{feat:Urls} are the numbers of hashtags and URL, respectively. Note some authors focus only on \textit{unique} hashtags and URL, i.e. they do not count the same hashtag or URL twice. It is also possible to compute the proportion of the user's tweets which contain at least one hashtag or URL \cite{Benevenuto2010,Chu2012}, or an average number of hashtags or URLs by tweet \cite{Pennacchiotti2011}, or the associated standard deviation \cite{Uddin2014}. User regularly tweeting URLs are likely to be information providers \cite{Java2007}, however spammers also behave like this \cite{Benevenuto2010}, so this feature alone is not sufficient to distinguish them. Spammers additionally tend to use shortened URLs to hide their actual malicious destination, or the fact the same URL is repeated many times \cite{Benevenuto2010,Wang2010}. Certain authors use blacklists of URLs in order to identify the tweets containing malicious ones \cite{Chu2012,Ghosh2012}.

In extreme cases, certain users like to fill their tweets with hashtags or URLs, much more than the regular users. For instance, certain social capitalists publish some tweets containing only hashtags, all related to the strategy they apply to gain visibility and exhort other people to subscribe to their feed (e.g. \texttt{\#TeamFollowBack}, \texttt{\#Follow4Follow}, cf. \cite{Dugue2014}).

Feature \ref{feat:AutoSimilarity} consists in processing the self-similarity of the user's tweets, i.e. the similarity between each pair of tweets he published, then using a statistic such as the average to summarize the results \cite{Lee2011,Lee2013}. Alternatively, one can also set a similarity threshold allowing to determine if two tweets are considered as similar, and count the pairs of similar tweets (or use some derived statistic) \cite{Wang2010}. This feature was notably used in studies aiming at detecting spammers: these users tend to post many times the same tweets, or very similar ones \cite{Lee2010,Lee2011,Wang2010}.

\subsubsection{External Data}
This category contains features corresponding to data not retrieved directly from Twitter. Feature \ref{feat:WebSearch} is simply the number of results returned by some Web search engine, which point at the user's Webpage. 

The next two features are scores computed by private companies independent from Twitter, and aim at measuring (in one way or another) the influence of users. Of course, they differ in the definition of the notion of influence they rely upon. Feature \ref{feat:Klout} is the Klout score, that takes into account both Tweeter-related and external data gathered from other social networking services and various search engines~\cite{KloutPaper}. The precise list of the features used to compute the Klout score was not published, though. The algorithm behind the Kred Influence Measurement \cite{Kred2015} is open source (Feature \ref{feat:Kred}). It is constituted of two scores: Influence (how the user's tweets are received by others) and Outreach (how much the user tend to spread other's tweets).

\subsection{General Remarks}
\label{sec:FeatGenRemarks}
We conclude our review with three remarks concerning all features. First, an important fact regarding the selection of features is their availability. Depending on the context of the considered study, all the features we listed cannot necessarily be used, for several reasons. First, the dataset given for the study might be incomplete, relatively to the features one wants to process. For instance, if one has access to a collection of Tweets, he still has to retrieve the subscription information to be able to use Features from category \textit{Local Connections}. But the Twitter API queries limitations might prevent him to access these data, or the concerned accounts may no longer exist, or the users may have changed their privacy settings. Some users also do not fill all the available fields, making it hard to use certain features from category \textit{User Profile}, unless the tool used to analyze the data is able to handle missing values. 

There are also time-related constraints: the data collected in practice only correspond to those that can be obtained in a reasonable amount of time. Moreover, even if one manages to retrieve all the necessary data, the computation of certain features can be very demanding if the dataset is too large, as we explained for Feature \ref{feat:InMentions}. Certain authors focus on the evolution of a given feature, by opposition to using a single value to summarize it. For instance, in \cite{Lee2011}, Lee \textit{et al}. measure the change rate of the number of followees (Feature \ref{feat:NbrFollow}). This can significantly complicate the data retrieval task, since this requires measuring the feature at different moments.

In our list, we omitted features one cannot compute in a normal context. For instance, when treating influence, Ramirez-de-la-Rosa \textit{et al}. use a feature corresponding to the type of job a user holds \cite{Ramirez2014}. However, this feature comes from the RepLab dataset (see Section \ref{sec:RepLab}) and was manually defined by a specialized agency. In practice, it is hardly possible to replicate exactly the same process on new data.

Our second remark concerns the way features are computed. We tried to stay general, and focus on what each feature represents conceptually. However, in practice, there are most of the times a number of ways to process a feature, which differ in various aspects. We indicated the main variants in the \textit{Descriptors} column of Table \ref{tab:Features}. However, we should emphasize that this aspect is much more important for content-related features, especially those from categories \textit{Lexical Aspects} and \textit{Stylistic Traits}. Indeed, those features coming from the NLP and IR fields are very sensitive to the way the content is pre-processed. The most common processes, such as removing punctuations (or emoticons and other special symbols as in \cite{Ramirez2014}) and hashtags marks, lower-casing the text, merging duplicated characters (i.e. turning \textit{whaaaat?} into \textit{what?}), can result in very different lexicon. Things get even more complicated when it comes to removing stop-words, since in practice each researcher uses his own list, often fitted manually to a specific issue.

Finally, it is worth noting certain authors define more complex features by combining basic ones, such as the ones we listed in Table \ref{tab:Features}. For instance, in \cite{Tommasel2015}, Tommasel \& Godoy define various ratios of the numbers of followers and followees (Feature \ref{feat:NbrFollow}), retweets (Features \ref{feat:OutRetweets} and \ref{feat:InRetweets}) and mentions (Features \ref{feat:OutMentions} and \ref{feat:InMentions}). In \cite{Lee2011}, Lee \textit{et al}. use the ratio of the total length of the mentions present in the tweet, to the overall tweet length, both expressed in characters. This amounts to dividing Feature \ref{feat:OutMentions} by Feature \ref{feat:TweetLength}. They also use the ratio of hashtag to tweet lengths, which is based on Features \ref{feat:Hashtags} and \ref{feat:TweetLength}.
Several other works use the same feature combination approach \cite{Anger2011,Benevenuto2010,Chu2012,Rao2010,Uddin2014,Wang2010}.

As mentioned before, the goal of this review was not to be exhaustive, which would be impossible given the number of works related to the characterization of Twitter users, but rather to present the most widespread and diverse features found in the literature. We focused on their meaning relatively to the user classification problem, and organized them in a new typology. As an illustration, in the rest of this article, we select some of these features and apply them to an actual problem: the prediction of \textit{offline influence}.

\section{Application to Offline Influence}
\label{sec:Application}
We illustrate the relevance of our feature review with an application to the prediction of offline influence based on Twitter data. In this section, we first define the notion of influence, and we discuss the difference between online and offline influence. We then describe RepLab 2014, a CLEF challenge aiming at the identification of Twitter users which are particularly influential in the real-world. Finally, we select a subset of the features presented in Section \ref{sec:Features}, in order to tackle this problem.

\subsection{Notion of Influence}
The \textit{Oxford Dictionary} defines influence as \textit{"The capacity to have an effect on the character, development, or behavior of someone or something"}. Various factors may be taken into account to measure the online influence of Twitter users. Intuitively, the more a user is followed, mentioned and retweeted, the more he seems influential \cite{Cha2010}. Nevertheless, there is no consensus regarding which features are the most relevant, or even if other features would be more discriminant. Most of the existing academic works consider the way the user is interacting with others (e.g. number of followers, mentions, etc.), the information available on his profile (age, user name, etc.) and the content he produces (number of tweets posted, textual nature of the tweets, etc). Several influence assessment tools were also proposed by companies such as Klout \cite{KloutPaper} and Kred \cite{Kred2015}.

Interestingly, these tools can be fooled by users implementing certain simple strategies. Messias \textit{et al}. \cite{Messias2013} showed that a \textit{bot} can easily appear as influential to Klout and Kred. Additionally, Danisch \textit{et al}. \cite{Danisch2014} observed that certain particular users called \textit{Social Capitalists} are also considered as influential although they do not produce any relevant content. Indeed, the strategy applied by social capitalists basically consists in following and retweeting massively each other. 
On a related note, Lee \textit{et al}. \cite{Lee2013} also showed that users they call \textit{Crowdturfers} use human-powered crowdsourcing to obtain retweets and followers. Finally, several data mining approaches were proposed regarding how to be retweeted or mentioned in order to gain visibility and influence \cite{Bakshy2011,Lee2014,Pramanik2015,Suh2010}.

A related question is to know how the user influence measured on Twitter (or some other online networking service) translates in terms of actual, real-world influence. In other words: how the \textit{online} influence matches the \textit{offline} influence. Some researchers proposed methods to detect \textit{Influencers} on the network, however except for some rare cases of very well known influential people, validation remains rarely possible. For this reason, there is only a limited number of studies linking real-life and network-based influence. Bond \textit{et al}. \cite{Bond2012} explored this question for Facebook, with their large-scale study about the influence of friends regarding elections, and especially abstention. They showed in particular that people who know that their Facebook friends voted are more likely to vote themselves. More recently, two conference tasks were proposed in order to investigate real-life influencers based on Twitter: PAN \cite{Rangel2014} and RepLab \cite{Amigo2014}. In this work, we focus on the latter, which is described in detail in the next subsection.

\subsection{RepLab Challenge}
\label{sec:RepLab}
The RepLab Challenge 2014 dataset \cite{Amigo2014} was designed for an influence ranking challenge organized in the context of the \textit{Conference and Labs of the Evaluation Forum}\footnote{\url{http://www.clef-initiative.eu/}} (CLEF). Based on the \textit{online} profiles and activity of a collection of Twitter users, the goal of this challenge was to rank these users in terms of \textit{offline} (i.e. real-world) influence. This is exactly the task we want to perform here, which makes this dataset particularly relevant to us. We therefore use these data for our own experiments. In this subsection, we first describe the context of the challenge and the data. Then, we explain how the performance was evaluated, and we discuss the results obtained during the challenge, as a reference for later comparisons. Finally, we present a classification variant of the problem, which we will tackle in addition to the ranking task.

\subsubsection{Data and task}
The main goal of the RepLab challenge is to detect \textit{offline} influence using \textit{online} Twitter data. The RepLab dataset contains users manually labeled by specialists from Llorente \& Cuenca\footnote{\url{http://www.llorenteycuenca.com/}}, a leading Spanish e-Reputation firm. These users were annotated according to their perceived real-world influence, and not by considering specifically their Twitter account,although annotators only considered users with at least $1,000$ followers. The annotation is binary: a user is either an \textit{Influencer} or a \textit{Not-Influencer}. The dataset contains a \textit{training set} of $2500$ users, including $796$ \textit{Influencers}, and a \textit{testing set} of $5900$ users, including $1563$ \textit{Influencers}. It also includes the $600$ last tweet IDs of each user, at the crawling and annotation time. This represents a total of $4,379,621$ tweets, i.e. around $750$ megabytes of data. These tweets can be written either in English or in Spanish. The dataset is publicly available\footnote{\url{http://nlp.uned.es/replab2014/}}.
RepLab finally provides a bounded and well designed framework to efficiency evaluate features and automatic influence detection systems.

Given the low number of real \textit{Influencers}, the RepLab organizers modeled the issue as a search problem restrained to the \textit{Automotive} and \textit{Banking} domains. In other words, the dataset was split in two, depending on the main activity domain of the considered users. The domains are mutually exclusive, i.e. one user belongs to exactly one domain. The objective was to rank the users in both domains in the decreasing order of influence. Both domains are balanced, with $2353$ (testing, including $764$ \textit{Influencers}) and $1186$ (training) users for the Automotive domain, and $2507$ (testing, $712$ \textit{Influencers}) and $1314$ (training) for the Banking domain. 

The organizers proposed a baseline consisting in ranking the users by descending number of followers. Basically, this consists in considering that the more a given user has followers, the more he is expected to be influential \textit{offline}. This baseline is directly inspired by \textit{online} influence measurement tools.

\subsubsection{Evaluation}
\label{sec:RepLabEval}
The RepLab framework \cite{Amigo2014} uses the traditional \textit{Mean Average Precision} (MAP) to evaluate the estimated rankings. The MAP allows comparing an ordered vector (output of a submitted method) to a binary reference (manually annotated data). In the case of RepLab, it was computed independently from the language, and separately for each domain.

For a given domain, the Mean Average Precision $MAP$ is computed as follows \cite{Buckley2000}:
\begin{equation}
\label{eq:map}
	MAP = \frac{1}{n} \sum_{i=1}^{N} P(i) q(i)
\end{equation}
\noindent where $N$ is the total number of users, $n$ the number of \textit{Influencers} correctly found (i.e. true positives), $P(i)$ the precision at rank $i$ (i.e. when considering the first $i$ users detected) and $q(i)$ is $1$ if the $i^{th}$ user is influential, and $0$ otherwise.

RepLab participants were compared according to the\textit{ Average MAP} processed over both \textit{Automotive} and \textit{Banking} domains.

\subsubsection{Results}
The UTDBRG group used Trending Topics Information, assuming that \textit{Influencers} tweet mainly about so-called \textit{Hot Topics} \cite{Aleahmad2014}. According to the official evaluation, their proposal obtained the highest MAP for the Automotive domain ($0.721$) and the best Average MAP among all participants ($0.565$). 
UAMCLYR combined user profile features and what they call \textit{writing behavior} (lexical richness, words and frequency of special characters) using Markov Random Fields \cite{Villatoro2014}. Still with an NLP perspective, ORM\_UNED \cite{MenaLomena2014} and LyS \cite{Vilares2014} investigated POS tags as additional features to those extracted from tweet contents. LyS also fed a classifier with bag-of-words built on the textual description published on certain profiles. Their proposal obtained the highest MAP for the Banking domain ($0.524$) and the second Average MAP among all participants ($0.563$).

Based on the assumption that \textit{Influencers} tend to use specific terms in their tweets, the LIA group opted to model each user based on the textual content associated to his tweets \cite{Cossu2014}. Using $k$-Nearest Neighbors ($k$-NN), they then matched each user to the most similar ones in the training set. More recently, the same team proposed some enhancements of this approach \cite{Cossu2015a}. They used a different tuning criterion and observed ranking improvements relatively to their official challenge submission which was outperformed with $0.764$ and $0.652$ MAP for Automotive and Banking, respectively, and a $0.708$ Average MAP. Also using a text-based method, our team (Cossu \textit{et al}. \cite{Cossu2015}) obtained even higher results with MAP reaching $0.803$ and $0.714$ for the Automotive domain and in Average, respectively. The performance for Banking remained lower with a $0.626$ MAP.

In RepLab participants submissions, performance differences observed between domains are likely due to the fact one domain is more difficult to process than the other. The \textit{Followers baseline} remains lower than most submitted systems, achieving a MAP of $0.370$ for Automotive and $0.385$ for Banking. All these values are summarized in Table \ref{tab:autrank}, in order to compare them with our own results.

\subsubsection{Classification Variant}
Because the reference itself is only binary, the RepLab ordering task can alternatively be seen as a binary classification problem, consisting in deciding if a user is an \textit{Influencer} or not. However, this was not a part of the original challenge. Ramirez \textit{et al}. \cite{Ramirez2014} recently proposed a method to tackle this issue. We will also consider this variant of the problem in the present article. 

To evaluate the classifier performance, Ramirez \textit{et al}. used the $F$-Score averaged over both classes, based on the Precision and Recall processed for each class, which is typical in categorization tasks. This \textit{Macro Averaged $F$-Score} is calculated as follows:
\begin{equation}
	F = \frac{1}{k} \sum\limits_{c} \dfrac{2 (P_c \times R_c)}{P_c + R_c}
\end{equation}

\noindent where $P_c$ and $R_c$ are the Precision and Recall obtained for class $c$, respectively, and $k$ is the number of classes (for us: $2$). The performance is considered for each domain (Banking and Automotive), as well as averaged over both domains. It gives an overview of the system ability to recover information from each class. 

Ramirez \textit{et al}. do not use any baseline to assess their results. Nevertheless, the imbalance between the influencer (31\%) and non-influencer (69\%) in the dataset leads to a strong non-informative baseline which simply consists in putting all users in the majority class (non-influencers). This baseline, called MF-Baseline (most frequent class baseline) achieves a $0.50$ Macro Averaged $F$-Score.

For this classification task, Ramirez \textit{et al}. reached a MAP of $0.696$ and $0.693$ for Automotive and Banking domains, respectively, and a $0.694$ Macro Averaged $F$-score. On the same task, our team (Cossu \textit{et al}. \cite{Cossu2015}) proposed a classification method based on tweet content, but obtained relatively low results ($0.40$ Macro Averaged $F$-Score).

\subsection{Experimental Setup}
\label{sec:ExperimentalSetup}
In order to tackle the offline influence problem, we adopted an exploratory approach: we do not know \textit{a priori} which features from Table \ref{tab:Features} are relevant for the considered problem. So, we selected as many of them as possible. However, we could not take advantage of all of them, or use all the descriptors available for a given feature, be it for computational or time issues, because the necessary data were not available, or simply for practical reasons. In this subsection, we list the selected features, which include both scalars and vectors. We also describe how we processed them, in function of their nature. The scripts\footnote{\label{foot:scripts}\url{https://github.com/CompNet/Influence}} corresponding to this processing are publicly available online, as well as the resulting outputs\footnote{\label{foot:outputs}\url{http://dx.doi.org/10.6084/m9.figshare.1506785}}.

\subsubsection{Scalar Features}
\label{sec:ExpSetupScalar}
We selected scalar features from each category of Table \ref{tab:Features}: \textit{User Profile} (Features \ref{feat:Picture}--\ref{feat:DescrLength}), \textit{Publishing Activity} (Features \ref{feat:Tweets}, \ref{feat:Delay} and \ref{feat:Geolocated}), \textit{Local Connections} (Features \ref{feat:NbrFollow}--\ref{feat:IdFollow}), \textit{User Interaction} (Features \ref{feat:OutRetweets}--\ref{feat:OutMentions}), \textit{Stylistic Traits} (Features \ref{feat:TweetLength}, \ref{feat:Hashtags} and \ref{feat:Urls}), and \textit{External Data} (Features \ref{feat:WebSearch} and \ref{feat:Klout}). For \textit{Lexical Aspects}, as explained in Section \ref{sec:ExpSetupCooc}, we defined additional scalar features by averaging several vectors corresponding to Feature \ref{feat:NgramWeights} (term cooccurrences or bigrams). 

Some of these features can be handled through several descriptors, so we had to make some additional choices. For Feature \ref{feat:Geolocated} (geolocation), we considered both the number of distinct locations from which the user twitted, and the proportion of geolocated tweets among his published tweets. Our intuition to consider geolocation-related features was that some users might tweet from some places of power or decision (relatively to their activity domain), which could be a good indicator of real-world influence. 
Regarding Feature \ref{feat:NbrFollow} (neighbors), we used the number of followers, number of followees, and the number of users which are both at the same time (i.e. cardinality of the intersection of the follower and followee sets). For Feature \ref{feat:IdFollow} (neighbors ids), we considered the standard deviation of the ids of the $5000$ most recent followers, and did the same for the followees. The topology of the follower-followee network has proven to be an important feature for the prediction of online influence, so it is worth a try when dealing with offline influence. 
We investigated Feature \ref{feat:TweetLength} (tweet length) considering average values expressed in terms of both number of characters and number of words. We discarded min and max values, because in our dataset they tend be the same ($0$ and $140$) for all users. We think tweet length is likely to be relevant to identify autorities, which we suppose have more to say than non-influential people.
For Feature \ref{feat:OutMentions} (mentions), we used the number of mentions by tweet, number of unique mentions by tweet, proportion of tweets that contain mentions, and total number of distinct usernames mentioned. Regarding \textit{Favorites} (Features \ref{feat:OutFavs} and \ref{feat:OutFavs}), we hypothesized that tweets from influential users are often marked as favorites by other users while influencers do not use this functionality. 
\todoVL{là je n'ai pas compris : on n'a pas utilisé les favoris, si ? En tout cas on ne l'avait pas indiqué dans la version soumise. Tu as donc corrigé une omission, JV ?}
For Feature \ref{feat:Hashtags} (hashtags), we used the number of unique hashtags, the number of hashtags by tweet, the number of unique hashtags by tweet, and the proportion of tweets that contain hashtags. We selected these features because previous results such as \cite{Aleahmad2014} indicate that user activity on trending topics is a great indicator of influence. 
Similarly, for Feature \ref{feat:Urls} (URLs), we distinguished the numbers of URLs by tweet, of unique URLs by tweet, and the proportion of tweets that contain URLs. Note that for the last 3 features, the uniqueness was determined over all the user's tweets (in the limit of the RepLab dataset), and not tweet-by-tweet. Our assumption here was that influential users tend to share links towards websites related with their profession or the activity domain, and possibly aiming at specific types of medias. However, for technical reasons, it was not possible to expend short URLs or to follow links, so we could not completely put this idea to the test.

We used non-linear classifiers under the form of kernelized SVMs (RBF, Polynomial and Sigmoid kernels) and logistic regression. We trained them using three distinct approaches: first with each scalar feature alone, second with all combinations of scalar features within each category defined by us (as described in Table \ref{tab:Features}, and third with all the scalar features at once. The two domains from the dataset (\textit{Banking} and \textit{Automotive}) were considered together and separately.

\subsubsection{Term Occurrences}
\label{sec:ExpSetupOcc}
As mentioned in Section \ref{sec:Features}, Feature \ref{feat:NgramWeights} focuses on the lexical aspect of tweets content. We now describe the different methods we used to take advantage of this feature. We focus on term occurrences, i.e. unigrams, in this subsection, and on term cooccurrences, i.e. bigrams, in the next. As a preprocessing step, the tweets were first lower-cased, we removed words composed of only one or two letters, URLs, as well as punctuation marks, but we kept mentions and hashtags as they were.

We defined our term-weighting using the classic \textit{Term Frequency -- Inverse Document Frequency} (TF-IDF) approach \cite{SparckJones1972}, combined with the \textit{Gini Purity Criterion} \cite{Gaussier2013}. We first introduce these measures in a generic way, before explaining how we applied them to our data.

The \textit{Term Frequency} $TF_d(i)$ corresponds to the number of occurrences of the term $i$ in the document $d$. The \textit{Inverse Document Frequency} $IDF(i)$ is defined as follows:
\begin{equation}
	\label{eq:idf} 
    IDF(i) = log(\frac{N}{DF(i)})
\end{equation}
\noindent where $N$ is the number of documents in the training set, and $DF(i)$ is the \textit{Document Frequency}, i.e. the number of documents containing term $i$ in the training set.

The purity $G(i)$ of a word $i$ is defined as follows: 
\begin{equation}
	\label{eq:gini} 
    G(i) = \sum_{c \in C} p(c|i)^2
    = \sum_{c \in C} \left ( \frac{DF_{c}(i)}{DF(i)} \right ) ^2
\end{equation}
\noindent where $C$ is the set of document classes and $DF_{c}(i)$ is the class-wise document frequency, i.e. the number of documents belonging to class $c$ and containing word $i$, in the training set. $G(i)$ indicates how much a term $i$ is spread over the different classes. It ranges from $1/|C|$ when a given word $i$ is well spread in all classes, to $1$ when the word only appears in a single class.

These measures are combined to define two distinct weights. First, the contribution $\omega_{i,d}$ of a term $i$ given a document $d$:
\begin{equation} 
	\label{eq:wid} 
	\omega_{i,d} = TF_d(i) \times IDF(i) \times G(i)
\end{equation}  

\noindent and second, the contribution $\omega_{i,c}$ of a term \textit{i} given a document class \textit{c}:
\begin{equation} 
	\label{eq:wic} 
	\omega_{i,c} = DF_c(i) \times IDF(i) \times G(i)
\end{equation}

Based on these weights, one can compute the similarity between a \textit{test} document $d$ and a document class $c$ using the \textit{Cosine} function as follows:
\begin{equation}
	\label{eq:cosinus}
	\cos(d,c) = \frac{\sum\limits_{i\in d \cap c} \omega_{i,d} \times \omega_{i,c}}{\sqrt{\sum\limits_{i\in d} \omega^{2} _{i,d} \times \sum\limits_{i\in c} \omega^{2} _{i,c}}}
\end{equation}
where $i$ represents a term, $d$ is the set of terms contained in the considered document, and $c$ is the set of all terms contained in the documents forming the considered class.

Now, let us see how we applied this generic approach to our specific case. First, note that each domain (Banking and Automotive) is treated separately, since a user belongs to only one domain. Regarding the languages (English and Spanish), we considered two approaches: processing all tweets at once without any regard for the language (called \textit{Joint} in the rest of the article) and treating the languages separately then combining the corresponding classes or ranking (\textit{Separated}). The process itself is two-stepped.

Our first step consists in determining which tweets to analyze for each user. We tested two different strategies: 1) use all the tweets provided by RepLab (strategy \textit{All}) ; and 2) select only the most relevant tweets (strategy \textit{Artex}). The latter consists in extracting only the 10\% most informative tweets the user published. For this purpose, we used a statistical Tweet Selection system developed in our research group, called \textit{Artex} \cite{Torres-Moreno2012}. Briefly, it relies on a $tf$--$idf$-based vector representation of, on one side, the user's average tweet, and on the other side, his vocabulary and sentences. The selection is performed by keeping tweets maximizing the cross-product between their vector, the vocabulary and the average tweet.

Our second step consists in classifying the users based on the Cosine similarity defined in Equation \ref{eq:cosinus}. We tested two distinct approaches, which are independent from the strategy used at the first step. In both approaches, the $i$ from Equation \ref{eq:cosinus} correspond to the terms remaining after our preprocessing, and the set $C$ contains two document classes, which are the two possible prediction outcomes: \textit{Influential} vs. \textit{Non-Influential}. However, the nature of the documents $d$ depends on the approach.

The first approach is called \textit{User-as-Document} (UaD) \cite{Kim2015}. It consists in merging all the tweets published by a user into a single large document. In other words, in this approach, a user is directly represented by a document $d$. A class is also represented by a single composite document, containing all the tweets written by the concerned users. For instance, the document representing the \textit{Influential} class is the concatenation of all tweets published by influential users. The classification process is performed by assigning a user to the most similar class, while the ranking depends on the similarity to the \textit{Influential} class. When the languages are treated separately (\textit{Separated} approach), we may obtain several different classes and rankings for each user, which need to be combined to get the final result. For this purpose, we weight the language-specific user-to-class similarities using the proportion of tweets belonging to the considered language, and sum. For instance, if the user posted twice as many English than Spanish tweets, the weight of the English similarity will be double of the Spanish one.

We call the second approach \textit{Bag-of-Tweets} (BoT), and it focuses on tweets instead of users. So this time, the documents $d$ from Equation \ref{eq:cosinus} correspond to tweets, and a user is represented by the set of tweets he published. A document class is also represented through such Bag-of-Tweets (i.e. influential vs. non-influential tweets). We compute the similarity between each user BoT and each class BoT, then decide the classification outcome using a voting process. We considered two variants: the first one (called \textit{Count}) consists in keeping the majority class among the user's tweets, whereas the second one (called \textit{Sum}) is based on the sum of the user's tweet similarity to the class \textit{Influencer}. The ranking is obtained by ordering users depending on the count or sum obtained for the Influential class. When the languages are treated separately (\textit{Separated} approach), document classes are represented by several distinct BoTs (one for each language). In order to combine the possibly different classes or rankings obtained for each language, we use the same approach than before: we weight the votes using the proportion of tweets belonging to the considered language.

\subsubsection{Term Cooccurrences}
\label{sec:ExpSetupCooc}
We also processed Feature \ref{feat:NgramWeights} based on bigrams. The tweets were preprocessed in the following way: the text was lowercased, we removed words with one or two letters, URLs, punctuation marks and stop-words (We used simple stop-lists available on the Oracle Website\footnote{\url{http://docs.oracle.com}}). Then, for each user, we processed a matrix representing how many times each word pair (bigram) appears consecutively, over all the tweets he posted. This consists in representing each user by a document containing all his tweets, like we did in the User-as-Document approach from the previous subsection, except the focus is now on coocurrences instead of occurrences. The obtained matrix is then considered as the adjacency matrix of the so-called cooccurrence graph. Each node in this graph represents a term, and the weight associated to a link connecting two nodes is the number of times the corresponding terms appear together in the text. 

Two users can be compared directly by computing the distance between their respective cooccurrence matrices. For this purpose, we simply used the Euclidean distance. We then applied the $k$ Nearest Neighbors method ($k$-NN) to separate Influential and Non-Influential users by matching each user of the test collection to the $k$ closest profiles of the training set. We tried different values of $k$, ranging from $1$ to $20$. During the voting process, each neighbor vote is weighted using his similarity to the user of interest. The ranking is obtained by processing a score corresponding to the sum of the influential neighbors' similarities. Like before, the domains were treated jointly and separately, and the results obtained for different languages are combined using the method previously described for the UaD approach(Section \ref{sec:ExpSetupOcc}).

It is also possible to summarize a cooccurrence graph through the use of a nodal topological measure, i.e. a function associating a numerical score to each node in the graph, describing its position in the said graph. Many such measures exist, taking various aspects of the graph topology into account \cite{FontouraCosta2007,Landherr2010}. We selected a set of classic nodal measures: \textit{Betweenness} \cite{Freeman1979}, \textit{Closeness} \cite{Bavelas1950}, \textit{Eigenvector} \cite{Bonacich1987} and \textit{Subgraph} \cite{Estrada2005} centralities, \textit{Eccentricity} \cite{Harary1969}, \textit{Local Transitivity} \cite{Watts1998}, \textit{Embeddedness} \cite{Lancichinetti2010}, \textit{Within-module Degree} and \textit{Participation Coefficient} \cite{Guimera2005}. These measures are described in detail in Appendix \ref{sec:CentralityMeasures}.
We selected them because they are complementary: certain are based on the \textit{local} topology (degree, transitivity), some are \textit{global} (betweenness, closeness, Eigenvector and subgraph centralities, eccentricity), and the others rely on the network community structure, and are therefore defined at an \textit{intermediary} level (embeddedness, within-module degree, participation coefficient).

Each nodal measure leads to a vector of values, each representing one specific term in the cooccurrence network. For a given measure, a user is consequently represented by such a vector. We process it using the same SVMs than for the scalar features (Section \ref{sec:ExpSetupScalar}). Note that for the scalar features, each value of the SVM input vector represents a distinct feature, whereas here it corresponds to the centrality measured for one term.
Alternatively, we also computed the arithmetic means of these vectors, for each nodal measure taken independently, and used them as scalar features, as indicated in Section \ref{sec:ExpSetupScalar}.

\section{Results and Discussions}
\label{sec:Results}
In this Section, we present the results we obtained on the RepLab dataset. We consider first the classification task, then the ranking one. 
Finally, we use a more visual approach to illustrate our discussion about the prediction of offline influence based on the features extracted from Twitter data.

\subsection{Classification}
\label{sec:ResultsClassif}
The kernelized SVMs we applied did not converge when considering scalar features, be it individually, by category, by combining categories and all together. We obtained the same behavior for the vector descriptors extracted from Feature \ref{feat:NgramWeights} (bigrams). This means the centrality measures used to characterize the coocurrence network were inefficient to find a non-linear separation of our two classes. Those results were confirmed by the logistic regressions: none of the trained classifiers performed better than the most-frequent class baseline (all user as non-influential). We also applied Random forests, which gave the same results. Meanwhile, as stated in Section \ref{sec:ExperimentalSetup}, these classifiers usually perform very well for this type of task.

However, we obtained some results for the remaining descriptors of Feature \ref{feat:NgramWeights}, as displayed in Table \ref{tab:autclas}. The classification performances are shown in terms of $F$-Score for each domain and averaged over domains, as explained in Section \ref{sec:RepLab}. For comparison purposes, we also reported in the same table the baseline, the results obtained by Ram{\'\i}rez-de-la-Rosa \textit{et al}. \cite{Ramirez2014} using SVM, and those of Cossu \textit{et al}. \cite{Cossu2015}, based on tweets content (Section \ref{sec:RepLab}).

\begin{table*}[!htb]
	\centering
	\tabcolsep = 2\tabcolsep
	\caption{Classification performances ordered by Average $F$-Score. \label{tab:autclas}}
    \begin{tabular}{l@{ }l@{ }l@{ }l@{ }p{3cm}lll}
        \hline
            \multicolumn{5}{l}{\textbf{Feature and descriptor}} & \textbf{Automotive} & \textbf{Banking} & \textbf{Average} \\
        \hline        
            Feature \ref{feat:NgramWeights} & User-as-Document & Separated & All & & .833 & .751 & .792 \\
            Feature \ref{feat:NgramWeights} & User-as-Document & Separated & Artex & & .829 & .745 & .787 \\            
            \multicolumn{5}{l}{Cossu \textit{et al} \cite{Cossu2015}} & .812 & .751 & .781 \\
            Feature \ref{feat:NgramWeights} & Bag-of-Tweets & Separated & Artex & Sum & .820 & .721 & .770 \\            
            Feature \ref{feat:NgramWeights} & Bag-of-Tweets & Separated & All & Sum & .817 & .709 & .763 \\            
            Feature \ref{feat:NgramWeights} & Bag-of-Tweets & Separated & Artex & Count & .796 & .719 & .757 \\
            Feature \ref{feat:NgramWeights} & Bag-of-Tweets & Separated & All & Count & .786 & .702 & .744 \\            
            Feature \ref{feat:NgramWeights} & User-as-Document & Joint & All & & .782 & .682 & .732 \\
            Feature \ref{feat:NgramWeights} & User-as-Document & Joint & Artex & & .773 & .672 & .722 \\
            \multicolumn{5}{l}{Ram{\'\i}rez-de-la-Rosa \textit{et al}. \cite{Ramirez2014}} & .696 & .693 & .694 \\
            Feature \ref{feat:NgramWeights} & Bag-of-Tweets & Joint & All & Count & .725 & .641 & .683 \\
            Feature \ref{feat:NgramWeights} & Bag-of-Tweets & Joint & All & Sum & .725 & .641 & .683 \\
			\multicolumn{5}{l}{\textit{MF-Baseline}} & .500 & .500 & .500 \\
			\multicolumn{5}{l}{Feature \ref{feat:NgramWeights} Cooccurrence networks} & .403 & .417 & .410 \\
		\hline
	\end{tabular}
\end{table*}

In Table \ref{tab:autclas}, one can observe that, except for the results provided by Ram{\'\i}rez-de-la-Rosa \textit{et al}. \cite{Ramirez2014}, the performance obtained for the \textit{Banking} domain is always lower than for the \textit{Automotive} domain. This confirms our observation from Section \ref{sec:RepLab}, regarding the higher difficulty to detect a user's influence for Banking than for Automotive.

As mentioned before, the cooccurrence networks extracted from Feature \ref{feat:NgramWeights} were processed by the $k$-NN method. The different $k$ values we tested did not lead to significantly different results, The best one is displayed in Table \ref{tab:autclas} and is clearly below the baseline for both domains. The features absent from the table were not able to reach the baseline level, let alone state-of-the-art scores.

The NLP cosine-based approaches applied to Feature \ref{feat:NgramWeights} showed competitive performances, noticeably higher than the baselines. Without language specific processing (\textit{Joint} method), the Bag-of-Tweets approach obtained state-of-the-art results, while the User-as-Document one outperformed all existing methods reported for this task, up to our knowledge. For both approaches, the performances are clearly improved when processing the languages separately (\textit{Separated} method). This might be due to the fact certain words are used in both languages, but in different ways. 

Regarding the decision strategy used for BoT, summing (\textit{Sum}) the votes improves the performance compared to simply counting them (\textit{Count}). This effect is more or less marked depending on the the way the languages are treated: no effect for \textit{Joint}, strong effect for \textit{Separated}. The domain also affects this improvement, which is much smaller for Banking than for Automotive. This could indicate users behave differently, in terms of how they redact tweets, depending on their domain. This would be consistent with our assumption regarding the use of different terminology by influential users of distinct activity domains.

The tweet selection step (approach \textit{Artex}) affects differently the BoT and UaD methods. For the former, there is an increase in performance, compared to using all available tweets (approach \textit{All}). Moreover, this increase is noticeably higher for Banking than for Automotive, which supports our previous observation regarding redactional differences between domains. The latter method (UaD), on the contrary, is negatively affected by \textit{Artex}. This can be explained in the following way: the tweet selection is a filter step, which reduces the noise contained in the user's Bag-of-Tweets, thus causing an increase in performance. However, the User-as-Document method already performs a relatively similar simplification, lexically speaking, so the improvement is much smaller, or can even turn into a deterioration.

The positive aspects of our results must be modulated by the fact the differences observed between the best unigram variants proposed for Feature \ref{feat:NgramWeights}, as well as Cossu \textit{et al}.'s method, are not statistically significant (according to a standard $t$-Test). More precisely, this observation concerns all rows from Table \ref{tab:autclas} between the first one and Ram{\'\i}rez-de-la-Rosa \textit{et al}.'s. The difference with Ram{\'\i}rez-de-la-Rosa \textit{et al}.'s method could not be tested directly, because we could not have access to their classification output. Our results nevertheless demonstrate that detecting offline influence is more efficiently tackled by taking content into account, rather than considering a large variety of text-independent features. In other words, for this task, writing similarities seem to be more relevant than any other Twitter-based information such as profile information, posting behavior or subscription-based interconnections.

\subsection{Ranking}
\label{sec:ResultsRanking}
The results obtained for the ranking task are displayed in Table \ref{tab:autrank} in terms of MAP, for each domain and averaged over domains. Again, one can observe that except for very few features, all scores are lower for the \textit{Banking} domain than for the \textit{Automotive} one.

\begin{table*}[!htb]
	\centering
	\tabcolsep = 2\tabcolsep
	\caption{Ranking performances ordered by Average MAP (the best ones are represented in bold).} 
    \label{tab:autrank}
	\begin{tabular}{l@{ }l@{ }l@{ }l@{ }p{3cm}lll}
    	\hline
        	\multicolumn{5}{l}{\textbf{Feature and descriptor}} & \textbf{Automotive} & \textbf{Banking} & \textbf{Average} \\
        \hline
           	Feature \ref{feat:NgramWeights} & User-as-Document & Separated & All & & \textbf{.803} & .626 & .\textbf{714} \\
			\multicolumn{5}{l}{Cossu \textit{et al}. \cite{Cossu2015}} & .764 & \textbf{.652} & .708 \\            
            Feature \ref{feat:NgramWeights} & Bag-of-Tweets & Separated & All & Sum & .779 & .628 & .703 \\
            Feature \ref{feat:NgramWeights} & Bag-of-Tweets & Separated & Artex & Sum & .774 & .633 & .703 \\
           	Feature \ref{feat:NgramWeights} & User-as-Document & Separated & Artex & & .782 & .623 & .702 \\
            Feature \ref{feat:NgramWeights} & Bag-of-Tweets & Separated & Artex & Count & .778 & .612 & .695 \\
            Feature \ref{feat:NgramWeights} & Bag-of-Tweets & Separated & All & Count & .762 & .592 & .677 \\
        	Feature \ref{feat:NgramWeights} & User-as-Document & Joint & All &  & .735 & .538 & .636 \\
            Feature \ref{feat:NgramWeights} & User-as-Document & Joint & Artex & & .722 & .547 & .634 \\
            Feature \ref{feat:NgramWeights} & Bag-of-Tweets & Joint & All & Sum & .699 & .526 & .612 \\
            Feature \ref{feat:NgramWeights} & Bag-of-Tweets & Joint & All & Count & .626 & .504 & .565 \\
            \multicolumn{5}{l}{UTDBRG -- Aleahmad \textit{et al}. \cite{Aleahmad2014}} & .721 & .410 & .565 \\
            \multicolumn{5}{l}{Feature \ref{feat:Tweets} Total Number of tweets} & .332 & .449 & .385 \\
            \multicolumn{5}{l}{Best Regression} & .424 & .338 & .381 \\
			\multicolumn{5}{l}{\textit{RepLab Baseline}} & .370 & .385 & .378 \\
        	\multicolumn{5}{l}{Feature \ref{feat:NgramWeights} Cooccurrence networks} & .298 & .300 & .299 \\
            \multicolumn{5}{l}{Feature \ref{feat:Klout} Klout score} & .304 & .275 & .289 \\
		\hline
	\end{tabular}
\end{table*}

The \textit{UTDBRG} row corresponds to the scores obtained at RepLab by the UTDBRG group \cite{Aleahmad2014}, which reached the highest average performance and the best MAP for \textit{Automotive}. This high performance for the Automotive domain, using an approach based on trending topics, probably reflects a tendency for Influencers to be up-to-date with the latest news relative to brand products and innovation in their domain. This statement is not valid for \textit{Banking}, where we can suppose that influence is based on more specialized and technical discussions. This is potentially why our previous approach (Cossu \textit{et al}.) based on tweets content obtained a good result for this domain, as mentioned in Section \ref{sec:RepLab}.

As mentioned in Section \ref{sec:ExpSetupScalar}, we first evaluated the logistic regression trained with each scalar feature alone, with each one of their categories, with each combination of category, and with all scalar features at once. The best results are presented on the row \textit{Best Regression}, and were obtained by combining the selected features of the following categories (cf. Table \ref{tab:Features}): \textit{User activity}, \textit{Profile fields}, \textit{Stylistic aspects} and \textit{External data}. The scores for this combination of features is just above the RepLab baseline, and far from the state-of-the-art approaches.

For each numerical scalar feature, we also considered the features values directly as a ranking method. The best results were obtained using the number of tweets posted by each user (Feature \ref{feat:Tweets}). Although its average MAP is just above the baseline, the performance obtained for the \textit{Banking} domain is above UTDBRG, the previous state-of-the-art results. Thus, we may consider this feature as the new baseline of this specific domain. All others similarly processed features remain lower than the official baseline. The results obtained for Feature \ref{feat:Klout} reflect very poor rankings. This is very surprising, because this feature is the Klout Score, which was precisely designed to measure influence in general (i.e. both on- and offline).

The rest of the results presented in Table \ref{tab:autrank} are the best we obtained for Feature \ref{feat:NgramWeights}. Those obtained using the direct comparison of cooccurrence networks are slightly better than for the Klout Score. The cosine-based methods applied to Feature \ref{feat:NgramWeights} led to very interesting results. The Bag-of-Tweets method obtained an average state-of-the-art performance, while the User-as-Document method reaches very high average MAP values, even larger than the state-of-the-art, be it domain-wise (for Automotive and Banking) or in average. 

Compared to the classification results, the performances of the BoT and UaD methods are tighter, but the latter still dominate the former, though. Again, both methods get better results when the languages are treated separately (approach \textit{Separated}). The BoT method still appear to perform better when using the \textit{Sum} decision strategy (instead of \textit{Count}). Including the tweet selection step (\textit{Artex}) showed no significant performance changes, be it in terms of increase or decrease. This means describing a user based on the vocabulary he uses over all his tweets retains the information necessary to rank his influence level.

Our results indicate that influential users from a specific domain behave differently and write in a particular manner compared to other users. In other words, \textit{Influencers} are characterized by a certain editorial behavior. For bilingual users, as observed for the classification task, separating their tweets in order to process the languages separately led to improvements in the ranking performance. This suggests that words originating from one language get a different meaning when used in the context of the other language.

Ram{\'\i}rez-de-la-Rosa \textit{et al}. \cite{Ramirez2014} were able to take advantage of certain scalar features to feed SVM-based classifiers in order to tackle the classification task, while RepLab participants such as the LyS \cite{Vilares2014} and UNED\_ORM \cite{MenaLomena2014} groups did the same for the ranking task. However, we were not able to obtain any results when using the same classification tools and similar features (no convergence). The large variety of descriptors that can be considered for each feature may explain this difference: a wrong descriptor choice is quite sufficient to mislead the training process. Yet, it is sometimes difficult or even impossible to find all the required details in the literature or the Web. This is the reason why we put our source code\textsuperscript{\ref{foot:scripts}} and outputs\textsuperscript{\ref{foot:outputs}} online, in order to ease the replication of the process which led to the results presented in this article. 

Despite this performance reproduction point, our NLP-based methods reached higher scores than state-of-the-art works, for both classification and ranking. This indicates that typical SNA features classically used to detect spammers, social capitalists or influential Twitter users, are not very relevant to detect \textit{offline Influencers}. In other terms, these typical features might be efficient to characterize influence perceived on Twitter, but not outside of it. Compared to other previous content-based methods, our approach consisting in representing a user under various forms of tweet bags-of-words also gave very good results. In particular, our User-as-Document method was far better than the best state-of-the-art approaches for both classification and ranking tasks. We suppose the way a user writes his tweets is related to his offline influence, at least for the studied domains. However, our attempt to extend this occurrence-based approach to a cooccurrence-based one using graph measures did not lead to good performances.


\subsection{PLS path modeling}
\label{sec:ResultsRegression}
In this last subsection, we come back to the scalar features and deepen the study of their relationship with offline influence through the use of \textit{Partial Least Squares Path Modeling} (PLS-PM) \cite{Wold1982}. 

The PLS algorithm handles all kinds of scales and is known to be well suited to combine nominal and binary variables. PLS-PM allows to represent a set of variables as a structure made of blocks of manifest (observed) variables. Each block is summarized by a latent variable, which depends on all the manifest variables constituting the block. PLS-PM estimates the best weights (between the manifest and latent variables, and between the latent variable and the predicted variable), by calculating the solution of the general underlying model of multivariate PLS \cite{Henseler2010}. The $R$ index is used to estimate the model quality (maximizes the square sum of correlations inside latent variables and between related variables). PLS-PM is a confirmatory approach which need an initial conceptual model derived from experts knowledge and also allows to extract information from the data. Furthermore, it offers a graphical representation of the relations between manifest and latent variables, which is valuable for analysis, even by non-specialists. For an extensive review and more details on PLS path modeling, see \cite{Tenenhaus2004}. 

Our application case (influence detection) can be viewed as a customer satisfaction index analysis as defined by Fornell \cite{Fornell1992}. We propose a conceptual model combining the predefined feature categories we defined Section \ref{sec:Features} (cf. Table \ref{tab:Features}). Our objective is to explain why classifiers exploiting these features failed, and to discover robust relations between latent variables. We also intend to investigate links between the features we selected and the values proposed by the best classifier applied to Feature \ref{feat:NgramWeights}, since it performed very well. Our model has $4$ hierarchical levels: first the features (manifest variables), each one connected to its category (latent variable), constituting the second level. Each category is in turn connected to either a \textit{Classifier} variable (representing the classifier output) or a \textit{Reference} variable (representing the ground truth from RepLab). We connected the content-based categories to Classifier (which is itself content-based), whereas the rest are connected to Reference. The Classifier variable is itself the third level, since it is also connected to the Reference variable the classifier output is supposed to be related to the actual influence). In other words, there are two types of categories in our model: those that directly induce \textit{Reference}, and those related to the classifier, which in turn induces the Reference.

\begin{figure}[!ht]
	\centering
    \textit{Automotive} Domain\\
 	\begin{minipage}{0.49\linewidth}
 		\centerline{\includegraphics[width=\linewidth]{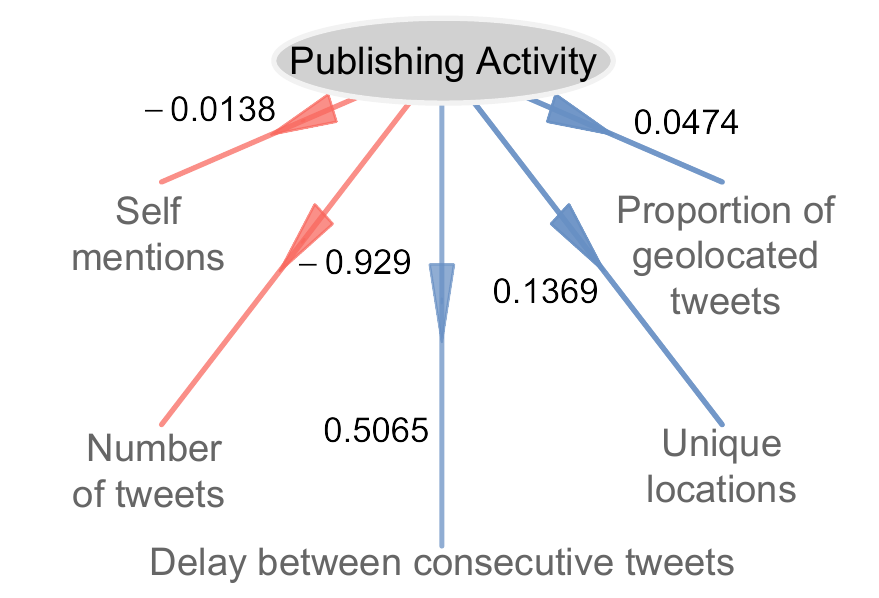}}
 	\end{minipage}
 	\begin{minipage}{0.49\linewidth}
 		\centerline{\includegraphics[width=\linewidth]{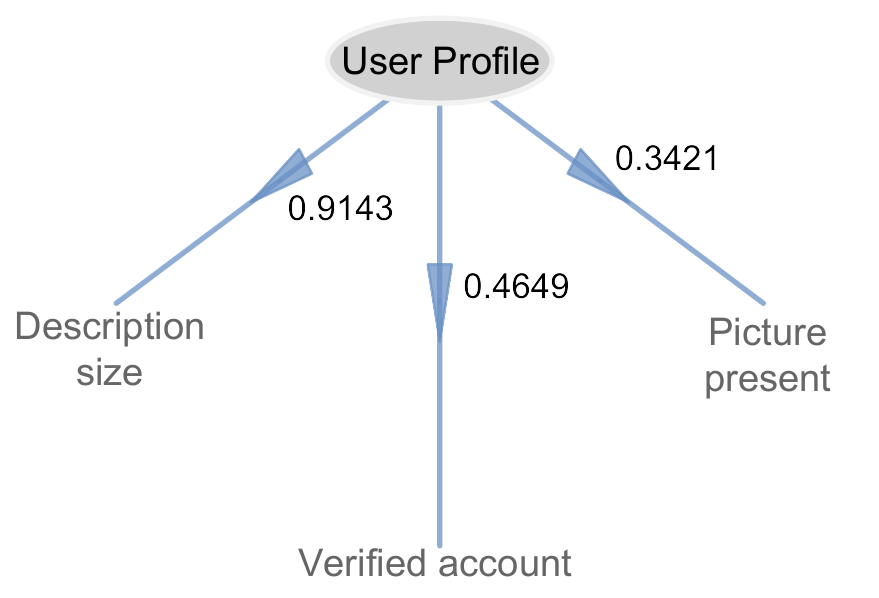}}
    \end{minipage}

    \vspace{10pt}\textit{Banking} Domain\\
	\begin{minipage}{0.49\linewidth}
 		\centerline{\includegraphics[width=\linewidth]{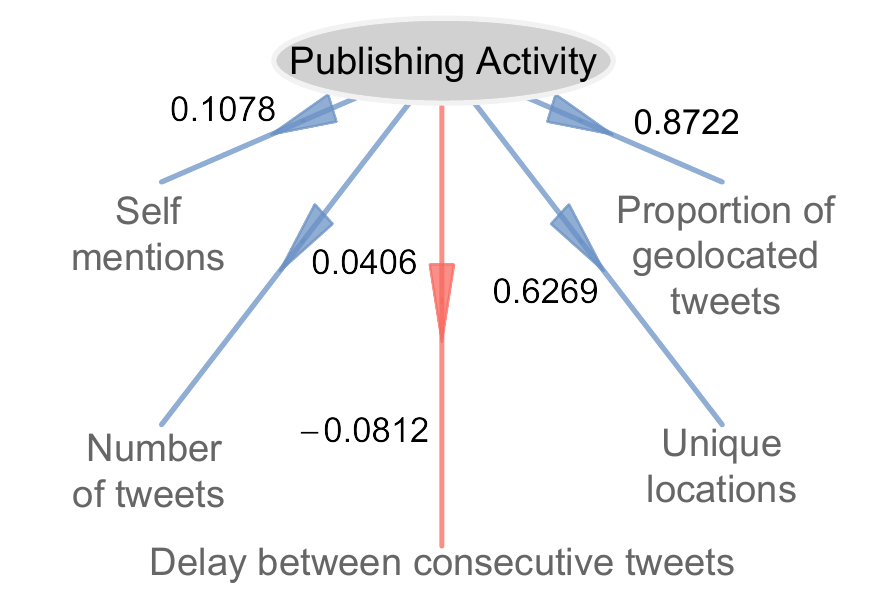}}
 	\end{minipage}
 	\begin{minipage}{0.49\linewidth}
 		\centerline{\includegraphics[width=\linewidth]{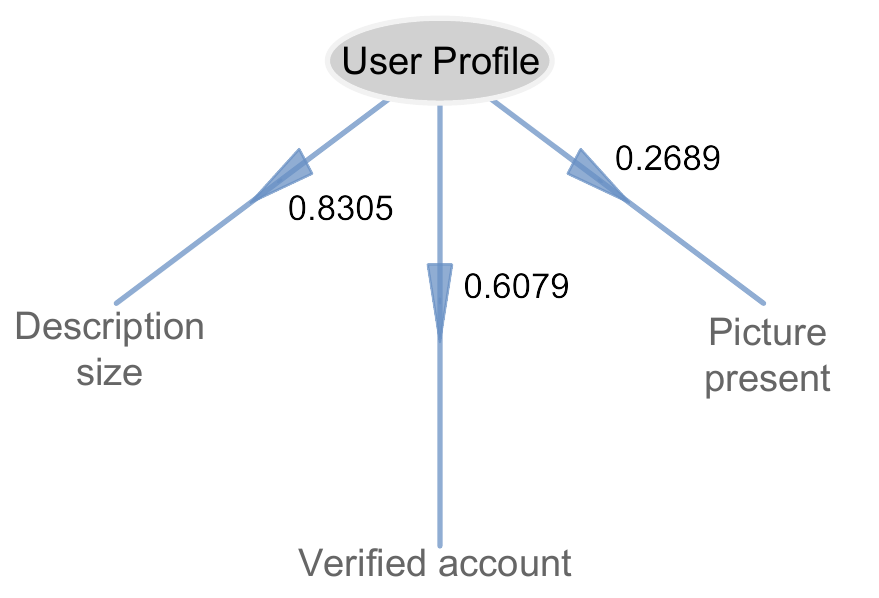}}
    \end{minipage}
 	\caption{Internal correlation of two latent variables: \textit{Publishing Activity} (left column) and \textit{User Profile} (right column) categories, for the \textit{Automotive} (top row) and \textit{Banking} (bottom row) domain models. \label{fig:acti_local}}
\end{figure}

The following experiments were made considering all users from the test set for which we could collect all features values, i.e. $2310$ and $2410$ users for Automotive and Banking, respectively. We selected as many features as possible and considered the method that obtained the best ranking result, that is to say: the \textit{UaD} method applied to \textit{All} tweets with \textit{Separated} languages. As an example, Figure \ref{fig:acti_local} shows the latent variables representing the \textit{Publishing Activity} and \textit{User Profile} categories, and their related manifest variables. The other categories are not displayed for space matters. The weights displayed in the figures correspond to the version of the correlation processed by PLS-PM. Note that a negative sign does not necessarily correspond to a negative correlation: PLS-PM select the signs in order to maximize the summed correlation values over the considered subgroup of variables.

Figure \ref{fig:acti_local} shows the features correlation differ depending on the domain. For the Automotive domain, Features \ref{feat:SelfMentions} and \ref{feat:Geolocated} have close to zero correlation values within their category, while Features \ref{feat:Delay} and \ref{feat:Tweets} reach much higher absolute values. Feature \ref{feat:Tweets}, in particular, is very close to $-1$, which is consistent with the observation we made in Section \ref{sec:ResultsRanking} regarding its use as a good baseline. For the Banking domain, it is quite the opposite: the Geolocation aspects are highly correlated, whereas the other features have a close to zero correlation. For \textit{User Profile}, the behavior is the same for both domains, with a strong correlation of Feature \ref{feat:DescrLength} (description length), and lesser correlation values for Features \ref{feat:Verified} (verified account) and \ref{feat:Picture} (image presence). It also indicates that influential users tend to have a complete account which allows people and mainly their followers to be sure about who their are.

We now describe quickly our results for the other categories (not represented here). For the \textit{Automotive} domain, the hashtag-related features are the main component of the \textit{Stylistic Traits} category. It confirms the intuition from Aleahmad \textit{et al}. \cite{Aleahmad2014} about the \textit{Influencers}' ability to be on the lookout for trending topics for this domain. For the \textit{Banking} domain, the numbers of URLs and Unique URLs obtained the highest scores in this category. According to this observation, future works should look toward computing an informativity index over both the tweets and the URLs they contain, in order to improve influence detection. Additional textual information from the targeted Web pages could also feed the NLP-based machine learning approaches to select the most relevant pages or part of pages. Concerning the \textit{Lexical Aspects} category, Feature \ref{feat:Words} (lexicon size) appears to be important for both domains, whereas Feature \ref{feat:Hapaxes} (hapaxes, i.e.  words specific to the user) reach a high correlation for the Automotive domain only.

\begin{figure}[ht]
    \centering
    \textit{Automotive} Domain\\
	\includegraphics[width=0.8\linewidth]{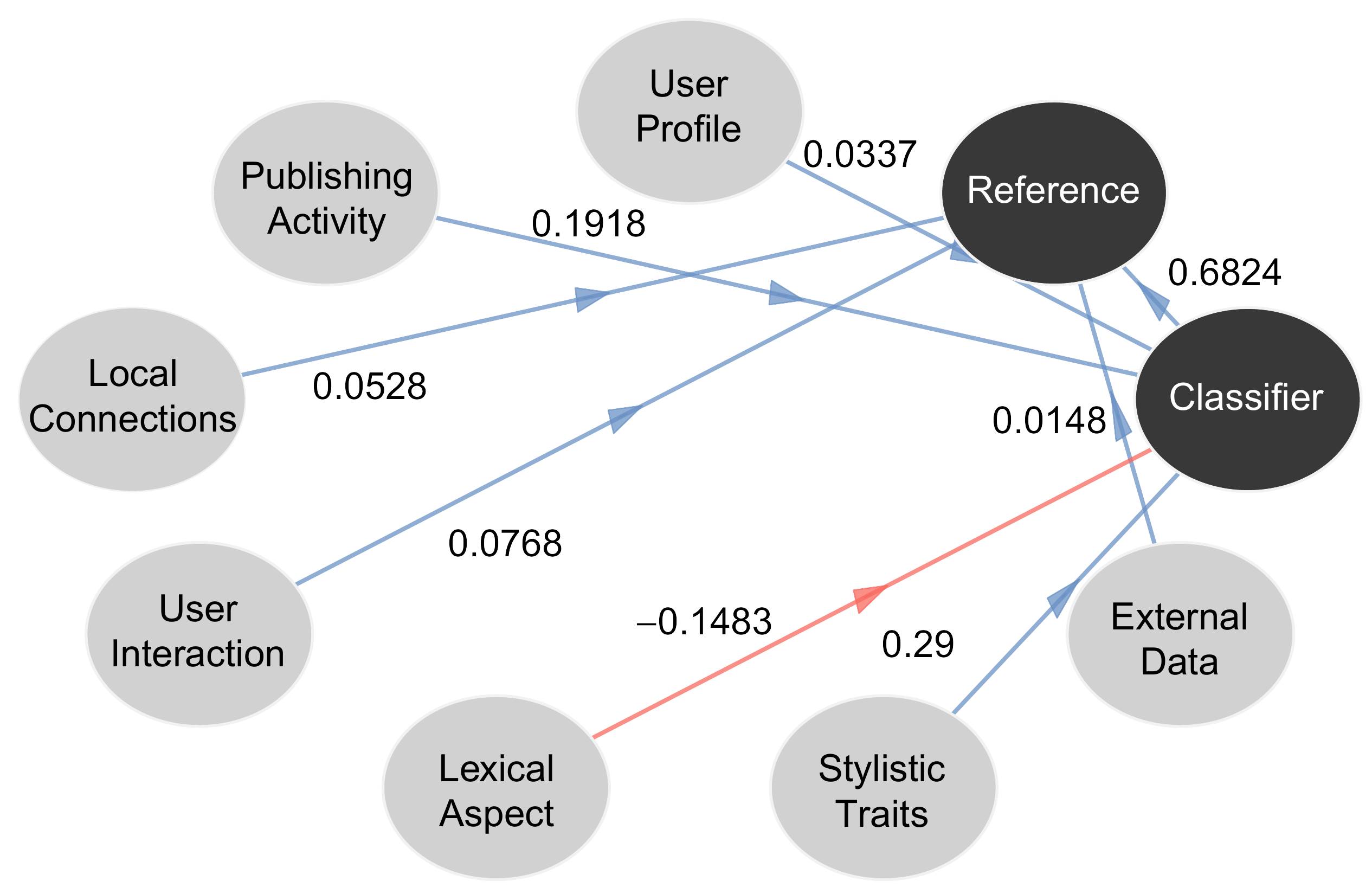}\\
    
    \textit{Banking} Domain\\
	\includegraphics[width=0.8\linewidth]{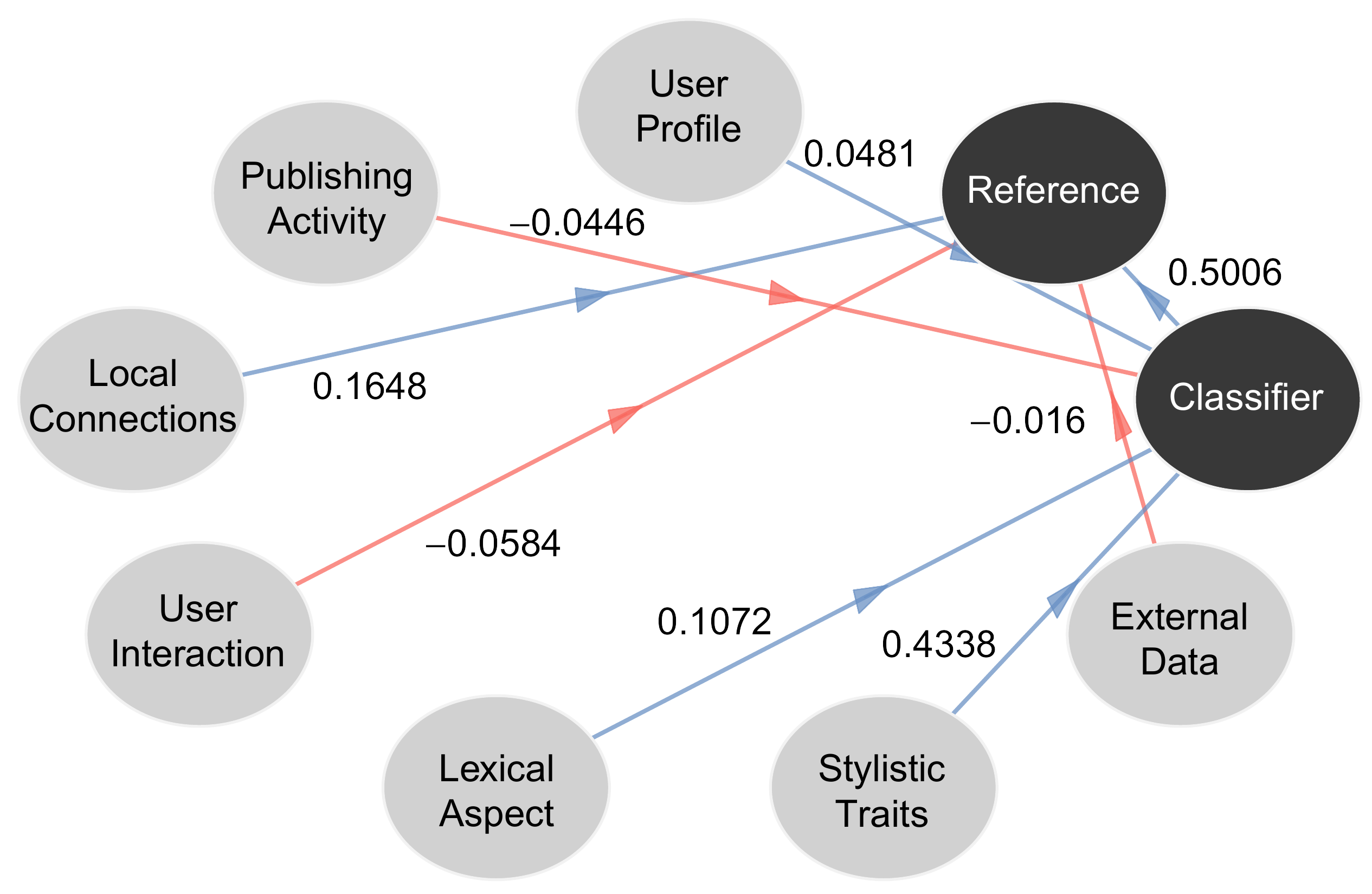}
	\caption{Correlation between the feature categories for the \textit{Automotive} (top) and \textit{Banking} (bottom) domains.
    \label{fig:external}}
\end{figure} 

Figure \ref{fig:external} depicts the second part of the regression model, i.e. the relationships between the latent variables and the Classifier and Reference variables, as well as the relationship between Classifier and Reference. The Classifier variable is clearly correlated to the Reference for both domains, although the values are closer to $0.5$ than $1$ which confirms the classification and ranking results obtained for Feature \ref{feat:NgramWeights}. Certain categories have close to zero correlation for both domain: \textit{User Profile}, \textit{User Interaction} and \textit{External Data} (which, in our case, contains only the Klout score) although the internal correlations within these categories are high. This means the categories are homogeneous, but not relevant for influence prediction. Some categories reach a larger than $0.1$ correlation (in absolute value): \textit{Publishing Activity} for Automotive, \textit{Local Connections} for Banking, and \textit{Lexical Aspects} for both. The differences observed between the domains confirm our assumption that the notion of offline influence takes a different form in Automotive and Banking. The \textit{Stylistic Traits} category has a much higher correlation than the other ones, for both domains, which highlights the interest of content-based features. Overall, the correlation between the categories and the Classifier and Reference variables is very low. This means the model is unable to find strong links with the influence estimation according to these latent variables, and can be related to the fact the SVMs did not converge when applied to these features.

\section{Conclusion}
\label{sec:Conclusion}
In this article, we have focused on the problem of user characterization on Twitter, and more particularly on the features used in the literature to perform such a classification. We first investigated a wide range of features coming from different research domains (mainly Social Network Analaysis, Natural Language processing and Information Retrieval), before proposing a new typology of features. 

We then tackled the problem of identifying and ranking real-life Influencers (a.k.a. \textit{offline} influencers) based on Twitter-related data, as specified by the RepLab 2014 challenge. For this experimental part, we can highlight two main results. First, we showed that classical SNA features used to detect spammers, social capitalists or users influential \textit{on Twitter}, do not give any relevant result on this problem. Our second result is the proposal of an NLP approach consisting in representing a user under various forms of bags-of-words, which led to a much better performance than all state-of-the-art methods (both content-based and -independent). From our result, we can suppose the way a user writes his tweets is related to his real-life influence, at least for the studied domains. This would confirm assumptions previously expressed in the literature regarding the fact users from specific domains behave and write in their own specific way.

It is important to highlight the fact our experimental results are valid only for the considered dataset. This means they are restricted to the domains it describes (Automotive and Banking), and are only as good as the manual annotation of the data. In RepLab 2014 \cite{Amigo2014}, the organizers were not able to conclude on significant differences between participants (and features or methods used) due to the small number of considered domains. Furthermore, the delay between our experiments and the annotation of the data may cause some bias, since certain users stopped their activities while others became more involved and earned followers. 

We think our results could be improved thanks to content-independent features. In particular, we hypothesize a more advanced use of the geolocation feature could help identifying geographical areas from which \textit{Influencers} tweet, e.g. financial places for the \textit{Banking} domain. Our approach based on cooccurrence graphs did not result in good performances, but could be improved in two ways. First, it is possible to use other graph measures, at different levels (micro, meso and macro) \cite{FontouraCosta2007}. Second, we could relax the notion of cooccurrence, by considering word neighborhoods of higher order.

\begin{acknowledgements}
This work is a revised and extended version of the article \textit{Detecting Real-World Influence Through Twitter}, presented at the $2^{nd}$ European Network Intelligence Conference (ENIC 2015) by the same authors \cite{Cossu2015}. It was partly funded by the French  National Research Agency (ANR), through the project ImagiWeb ANR-2012-CORD-002-01.
\end{acknowledgements}


\bibliographystyle{spmpsci}      
\bibliography{bibliography}

\appendix
\section{Centrality measures}
\label{sec:CentralityMeasures}
In their description, we note $G=(V,E)$ the considered cooccurrence graph, where $V$ and $E$ are its sets of nodes and links, respectively. 

The \textit{Degree} measure $d(u)$ is quite straightforward: it is the number of links attached to a node $u$. So in our case, it can be interpreted as the number of words co-occurring with the word of interest. More formally, we note $N(u)=\{v\in V:\{u,v\}\in E \}$ the \textit{neighborhood} of node $u$, i.e. the set of nodes connected to $u$ in $G$. The degree $d(u)=|N(u)|$ of a node $u$ is the cardinality of its neighborhood, i.e. its number of neighbors. 

The \textit{Betweenness} centrality $C_b(u)$ measures how much a node $u$ lies on the shortest paths connecting other nodes. It is a measure of accessibility~\cite{Freeman1979}:

\begin{equation}
	C_b(u) = \sum_{v < w}\frac{\sigma_{vw}(u)}{\sigma_{vw}} 
	\label{eq:betweenness}
\end{equation}

\noindent Where $\sigma_{vw}$ is the total number of shortest paths from node $v$ to node $w$, and $\sigma_{vw}(u)$ is the number of shortest paths from $v$ to $w$ running through node $u$.

The \textit{Closeness} centrality $C_c(u)$ quantifies how near a node $u$ is to the rest of the network \cite{Bavelas1950}:

\begin{equation}
	C_c(u) = \frac{1}{\sum_{v \in V} dist(u,v)}
	\label{eq:closeness} 
\end{equation}

\noindent Where $dist(u,v)$ is the \textit{geodesic distance} between nodes $u$ and $v$, i.e. the length of the shortest path between these nodes.

The \textit{Eigenvector} centrality $C_e(u)$ measures the influence of a node $u$ in the network based on the spectrum of its adjacency matrix. The Eigenvector centrality of each node is proportional to the sum of the centrality of its neighbors \cite{Bonacich1987}: 

\begin{equation}
	C_e(u) = \frac{1}{\lambda}\sum_{v \in N(u)}C_e(v)
	\label{eq:eigenvector} 
\end{equation}

\noindent Here, $\lambda$ is the largest Eigenvalue of the graph adjacency matrix.

The \textit{Subgraph} centrality $C_s(u)$ is based on the number of closed walks containing a node $u$ \cite{Estrada2005}. Closed walks are used here as proxies to represent subgraphs (both cyclic and acyclic) of a certain size. When computing the centrality, each walk is given a weight which gets exponentially smaller as a function of its length.

\begin{equation}
	C_s(u) = \sum_{\ell=0}^{\infty}\frac{\left(A^\ell\right)_{uu}}{\ell !}
	\label{eq:subgraph} 
\end{equation}

\noindent Where $A$ is the adjacency matrix of $G$, and therefore $\left(A^\ell\right)_{uu}$ corresponds to the number of closed walks containing $u$.

The \textit{Eccentricity} $E(u)$ of a node $u$ is its furthest (geodesic) distance to any other node in the network \cite{Harary1969}:

\begin{equation}
	E(u) = \max_{v \in V}(dist(u,v))
	\label{eq:eccentricity} 
\end{equation}

The \textit{Local Transitivity} $T(u)$ of a node $u$ is obtained by dividing the number of links existing among its neighbors, by the maximal number of links that could exist if all of them were connected \cite{Watts1998}:

\begin{equation}
	T(u) = \dfrac{|\{\{v,w\}\in E: v \in N(u) \wedge w \in N(u)\}|}{d(u)(d(u)-1)/2}
	\label{eq:transitivity}
\end{equation}

\noindent Where the denominator corresponds to the binomial coefficient $\binom{d(u)}{2}$. This measure ranges from $0$ (no connected neighbors) to $1$ (all neighbors are connected).

The \textit{Embeddedness} $e(u)$ represents the proportion of neighbors of a node $u$ belonging to its own community \cite{Lancichinetti2010}. The community structure of a network corresponds to a partition of its node set, defined in such a way that a maximum of links are located \textit{inside} the parts while a minimum of them lie \textit{between} the parts. We note $c(u)$ the community of node $u$, i.e. the parts that contains $u$.
Based on this, we can define the \textit{internal neighborhood} of a node $u$ as the subset of its neighborhood located in its own community: $N^{int}(u)=N(u) \cap c(u)$. Then, the \textit{internal degree} $d^{int}(u)=|N^{int}(u)|$ is defined as the cardinality of the internal neighborhood, i.e. the number of neighbors the node $u$ has in its own community.
Finally, the embeddedness is the following ratio:

\begin{equation}
	e(v) = \frac{ d_{int}(v)}{d(v)}
	\label{eq:embeddedness} 
\end{equation}

\noindent It ranges from $0$ (no neighbors in the node community) to $1$ (all neighbors in the node community).

The two last measures were proposed by Guimer\`a \& Amaral \cite{Guimera2005} to characterize the community role of nodes. For a node $u$, the \textit{Within Module Degree} $z(u)$ is defined as the $z$-score of the internal degree, processed relatively to its community $c(u)$:

\begin{equation}
	z(u) = \frac{ d_{int}(u)-\mu(d_{int},c(u))}{\sigma(d_{int},c(u))} 
	\label{eq:withinmoduledegree} 
\end{equation}

\noindent Where $\mu$ and $\sigma$ denote the mean and standard deviation of $d_{int}$ over all nodes belonging to the community of $u$, respectively. This measure expresses how much a node is connected to other nodes in its community, relatively to this community. By comparison, the embeddedness is not normalized in function of the community, but of the node degree.

The \textit{Participation Coefficient} is based on the notion of community degree, which is a generalization of the internal degree: $d_{i}(u)=|N(u) \cap C_{i}|$. This degree $d_{c}$ corresponds to the number of links a node $u$ has with nodes belonging to community number $i$. The participation coefficient is defined as:

\begin{equation}
	P(u) = 1-\sum_{1 \leq i \leq k} \left(\frac{d_i(u)}{d(u)}\right)^{2}
	\label{eq:participationcoeff} 
\end{equation}

\noindent Where $k$ is the number of communities, i.e. the number of parts in the partition. $P$ characterizes the distribution of the neighbors of a node over the community structure. More precisely, it measures the heterogeneity of this distribution: it gets close to $1$ if all the neighbors are uniformly distributed among all the communities, and to $0$ if they are all gathered in the same community. 

Both community role measures are defined independently from the method used for community detection (provided it identifies mutually exclusive communities). In this work, we applied the InfoMap method \cite{Rosvall2008}, which was deemed very efficient in previous studies \cite{Orman2012a}.

\end{document}